\RequirePackage{fix-cm}
\documentclass[twocolumn]{svjour3}          
\smartqed  
\usepackage[utf8]{inputenc}
\usepackage{graphicx}
\usepackage{subfig}
\usepackage{tabularx}
\usepackage{booktabs}
\usepackage[hyphens]{url}
\usepackage{siunitx}
\usepackage[misc]{ifsym}
\usepackage{microtype}
\usepackage{amsmath}
\newcommand{\fone}{F\textsubscript{1}}
\begin{document}
\title{Sequence-aware multimodal page classification of Brazilian legal documents
\thanks{This preprint, which was originally written on 8 April 2021, has not undergone peer review or any post-submission improvements or corrections. The Version of Record of this
article is published in the International Journal on Document Analysis and Recognition (IJDAR), and is available online at https://doi.org/10.1007/s10032-022-00406-7. The full paper can be accessed at https://rdcu.be/cRvvV.}
}


\author{Pedro H. Luz de Araujo         \and
        Ana Paula G. S. de Almeida \and
        Fabricio~A.~Braz \and
        Nilton C. da Silva \and
        Flavio de Barros Vidal \and
        Teofilo E. de~Campos 
}


\institute{P. H. Luz de Araujo (\Letter) $\cdot$ F. B. Vidal $\cdot$ T. E. de Campos \at
             Department of Computer Science, Universidade de Brasília, 70910-900 Brasília, Brazil \\
             \email{pedrohluzaraujo@gmail.com} 
             \\
             \email{fbvidal@unb.br} \\
             \email{t.decampos@oxfordalumni.org}
           \and
          A. P. G. S. de Almeida \at
          Department of Mechanical Engineering, Universidade de Brasília, 70910-900 Brasília, Brazil \\
          \email{anapaula.gsa@gmail.com}
           \and
           F. A. Braz $\cdot$ N. C. da Silva \at
             Gama Faculty, Universidade de Brasília, Campus Gama - Setor Leste -  Gama - 72444-240, Brasília, Brazil\\
              \email{\{fabraz, niltoncs\}@unb.br}           
}


\maketitle

\begin{abstract}
The Brazilian Supreme Court receives tens of thousands of cases each semester. Court employees spend thousands of hours to execute the initial analysis and classification of those cases---which takes effort away from posterior, more complex stages of the case management workflow. In this paper, we explore multimodal classification of documents from Brazil's Supreme Court. We train and evaluate our methods on a novel multimodal dataset of 6,510 lawsuits (339,478 pages) with manual annotation assigning each page to one of six classes. Each lawsuit is an ordered sequence of pages, which are stored both as an image and as a corresponding text extracted through optical character recognition. We first train two unimodal classifiers: a ResNet pre-trained on ImageNet is fine-tuned on the images, and a convolutional network with filters of multiple kernel sizes is trained from scratch on document texts. We use them as extractors of visual and textual features, which are then combined through our proposed Fusion Module. Our Fusion Module can handle missing textual or visual input by using learned embeddings for missing data. Moreover, we experiment with bi-directional Long Short-Term Memory (biLSTM) networks and linear-chain conditional random fields to model the sequential nature of the pages. The multimodal approaches outperform both textual and visual classifiers, especially when leveraging the sequential nature of the pages.

\end{abstract}

\section{Introduction}
\label{sec:intro}

The Brazilian court system is burdened by a large number of lawsuits. In 2019, there were 77.1 million lawsuits awaiting judgment---almost one lawsuit for every three Brazilians. Some of these lawsuits will stay in the system for a long time, with average processing times that can reach more than six years. All of this contributes to raising the legal system cost: that same year, Brazil spent about R\$100 billion in expenses with the judiciary, about 25 billion dollars considering the average exchange rate in 2019~\cite{justiceSummary2020}.

\begin{figure*}
  \setkeys{Gin}{width=\linewidth}
  \begin{tabularx}{\textwidth}{XXXX}
  \includegraphics{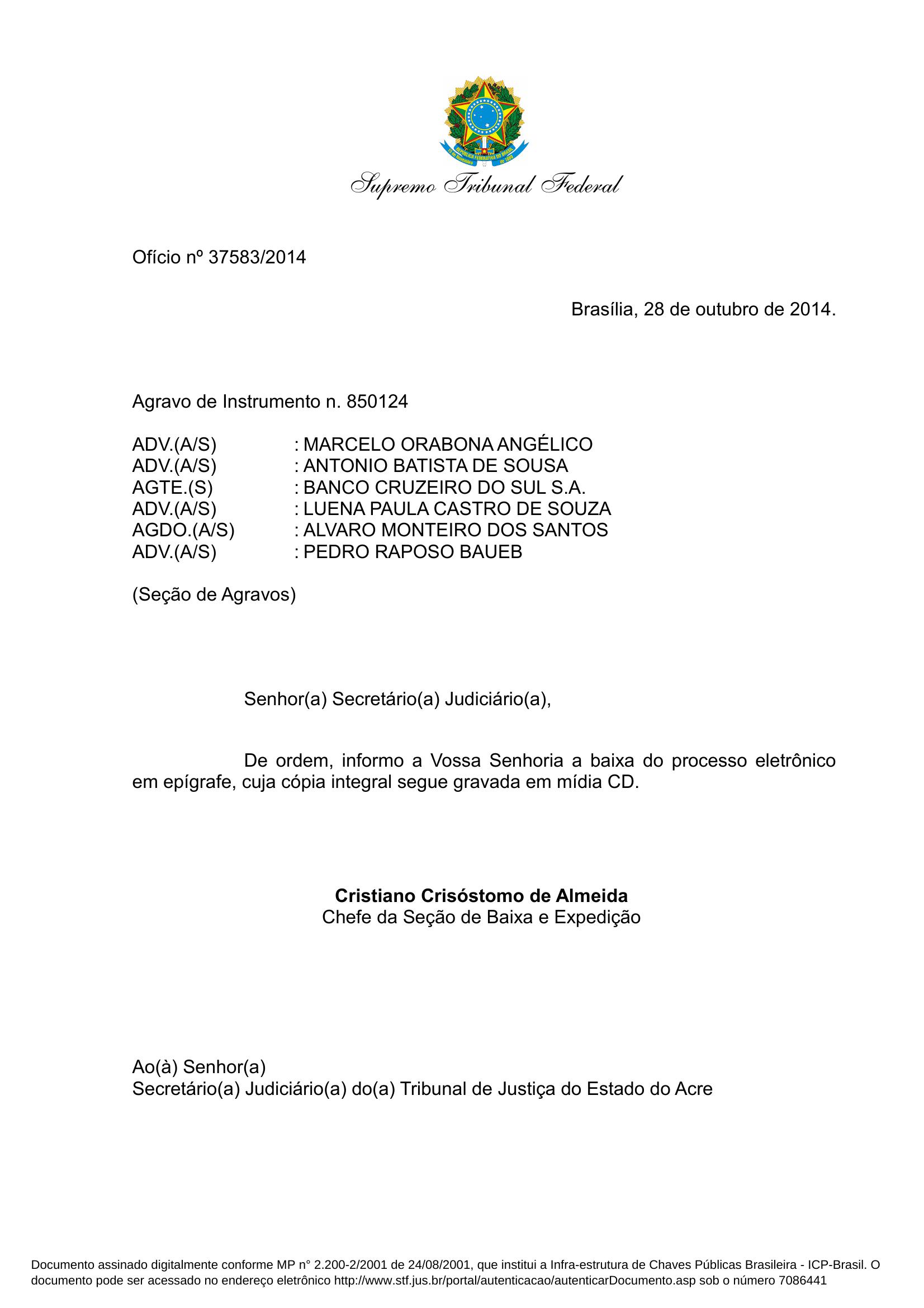}&
  \includegraphics{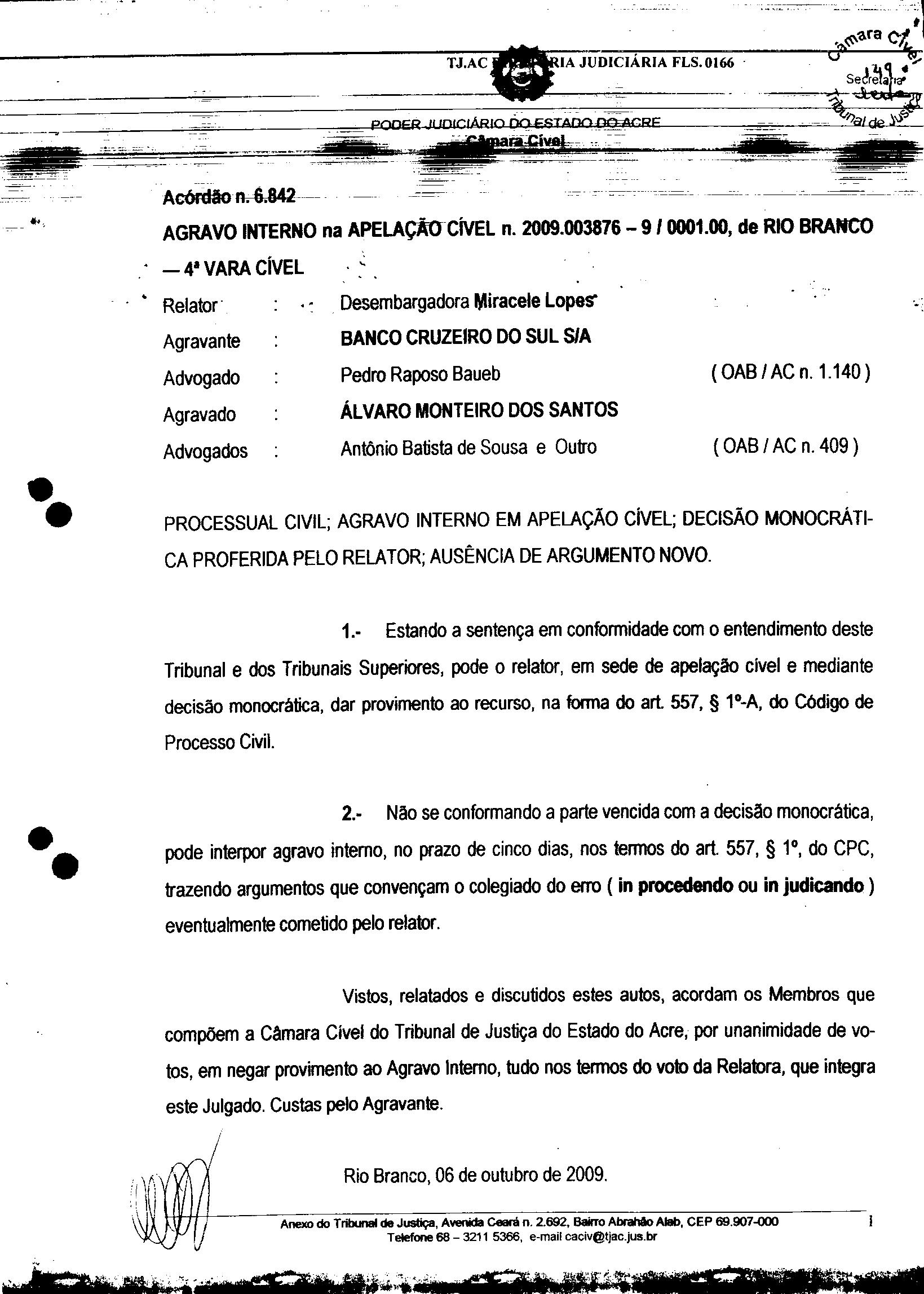}&
  \includegraphics{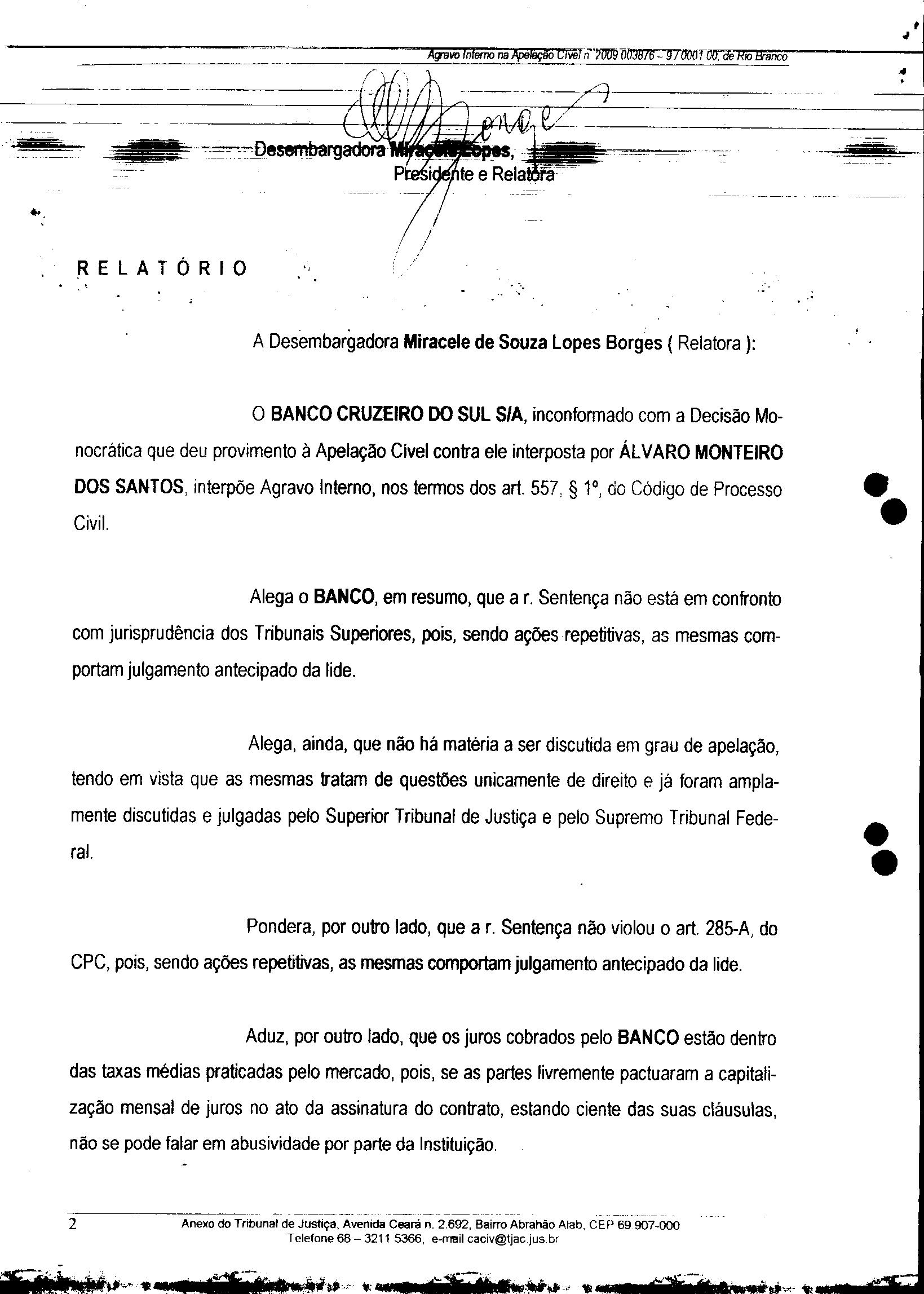}
  &\includegraphics{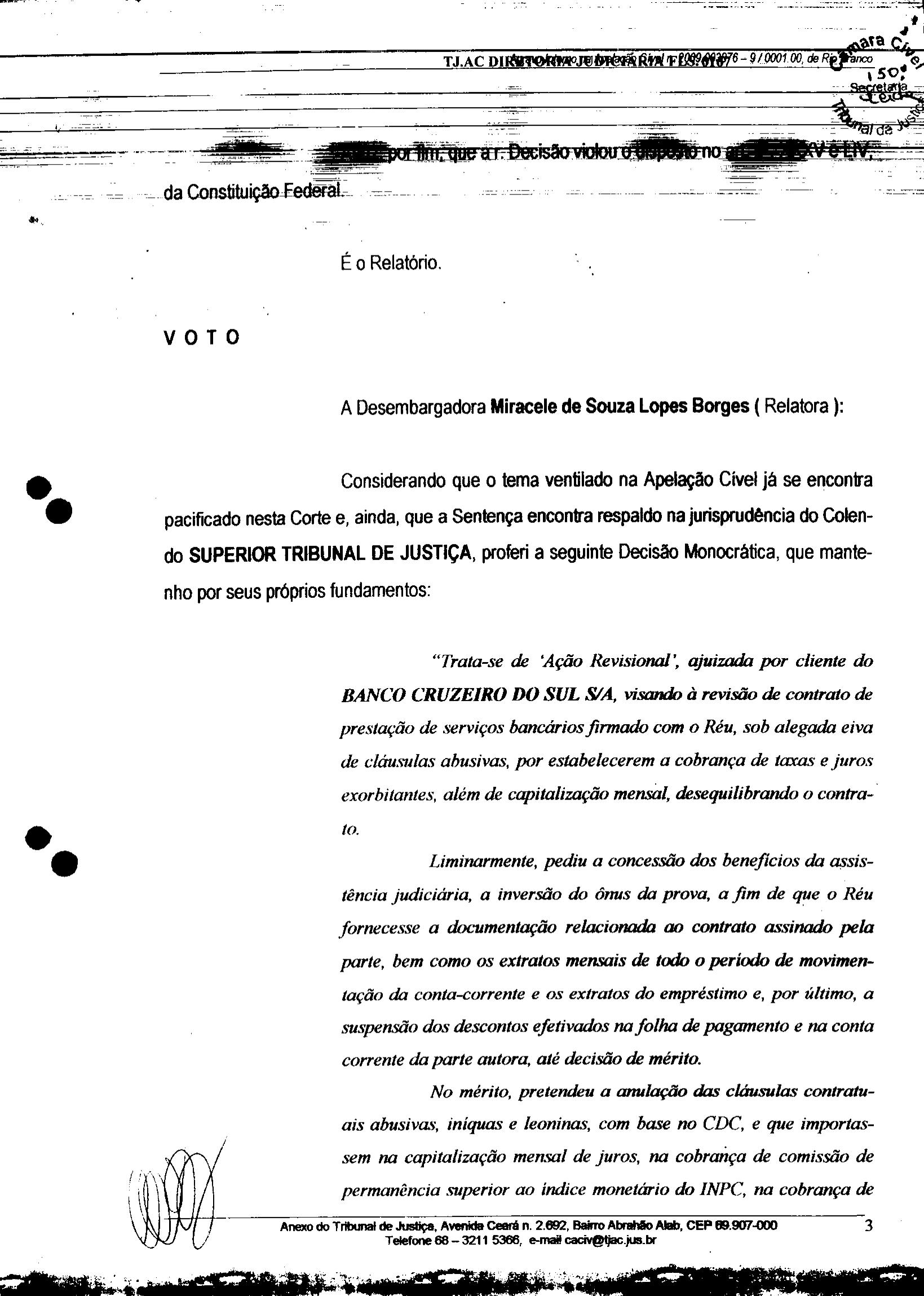}\\
  \includegraphics{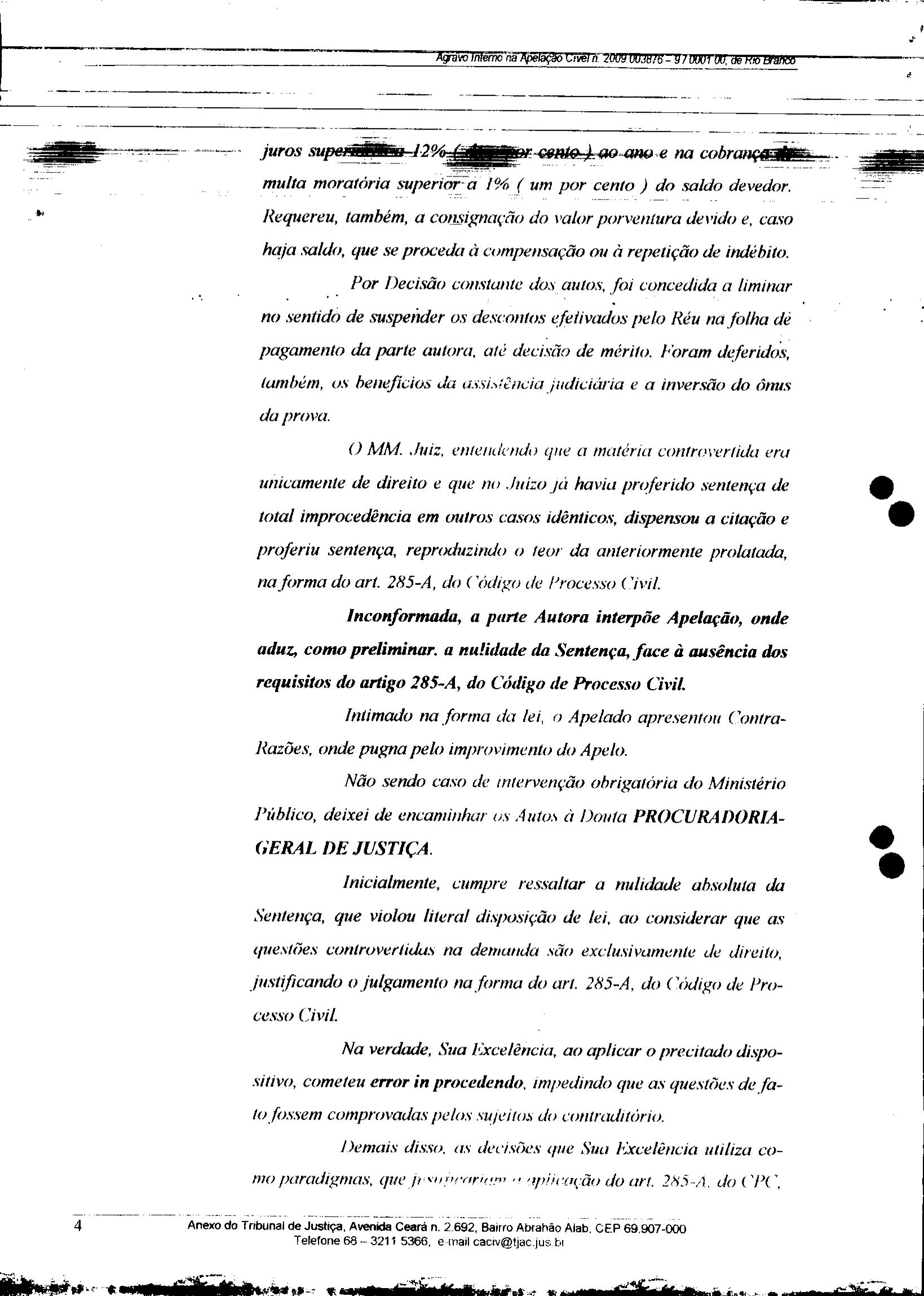}&
  \includegraphics{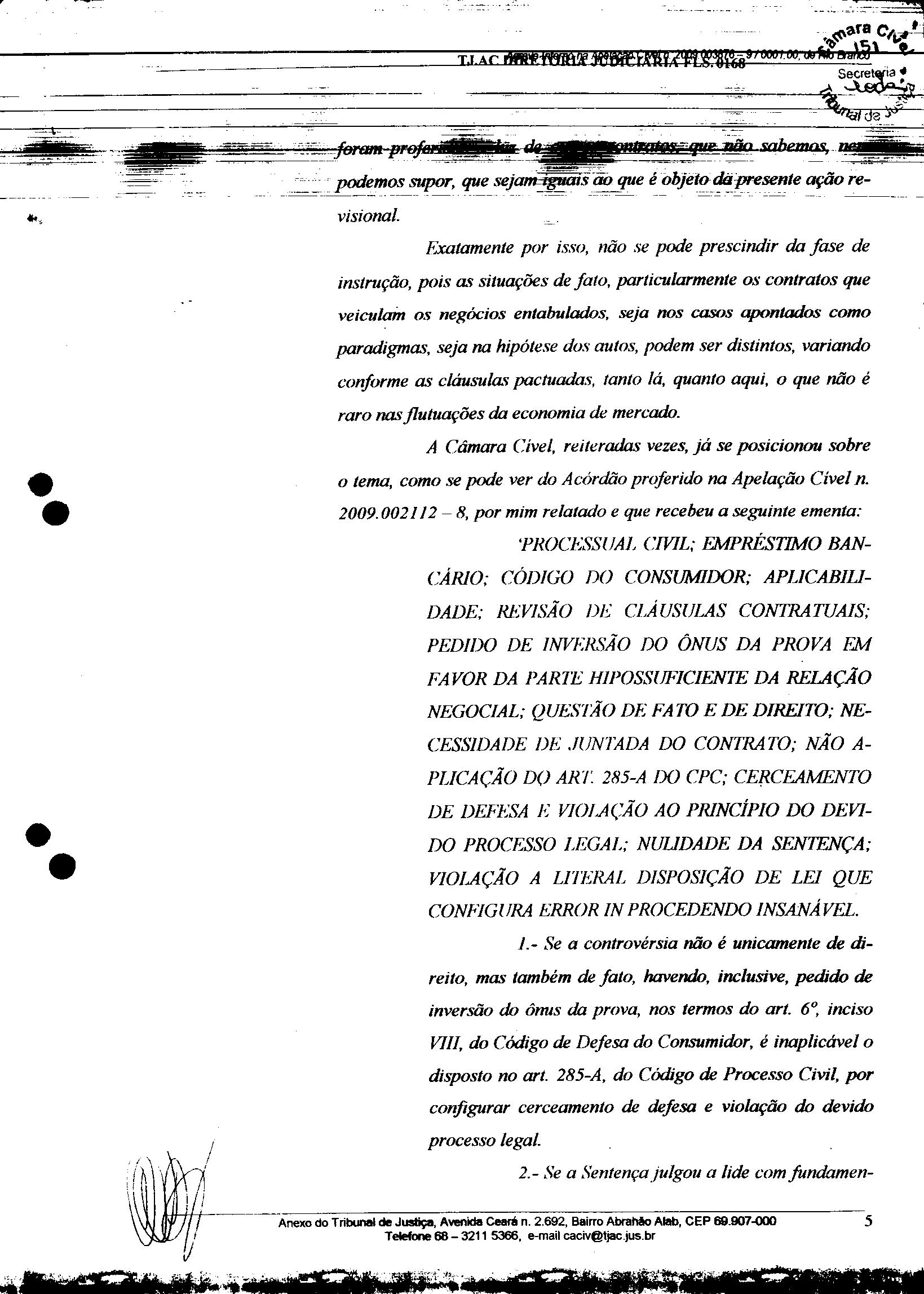}&
  \includegraphics{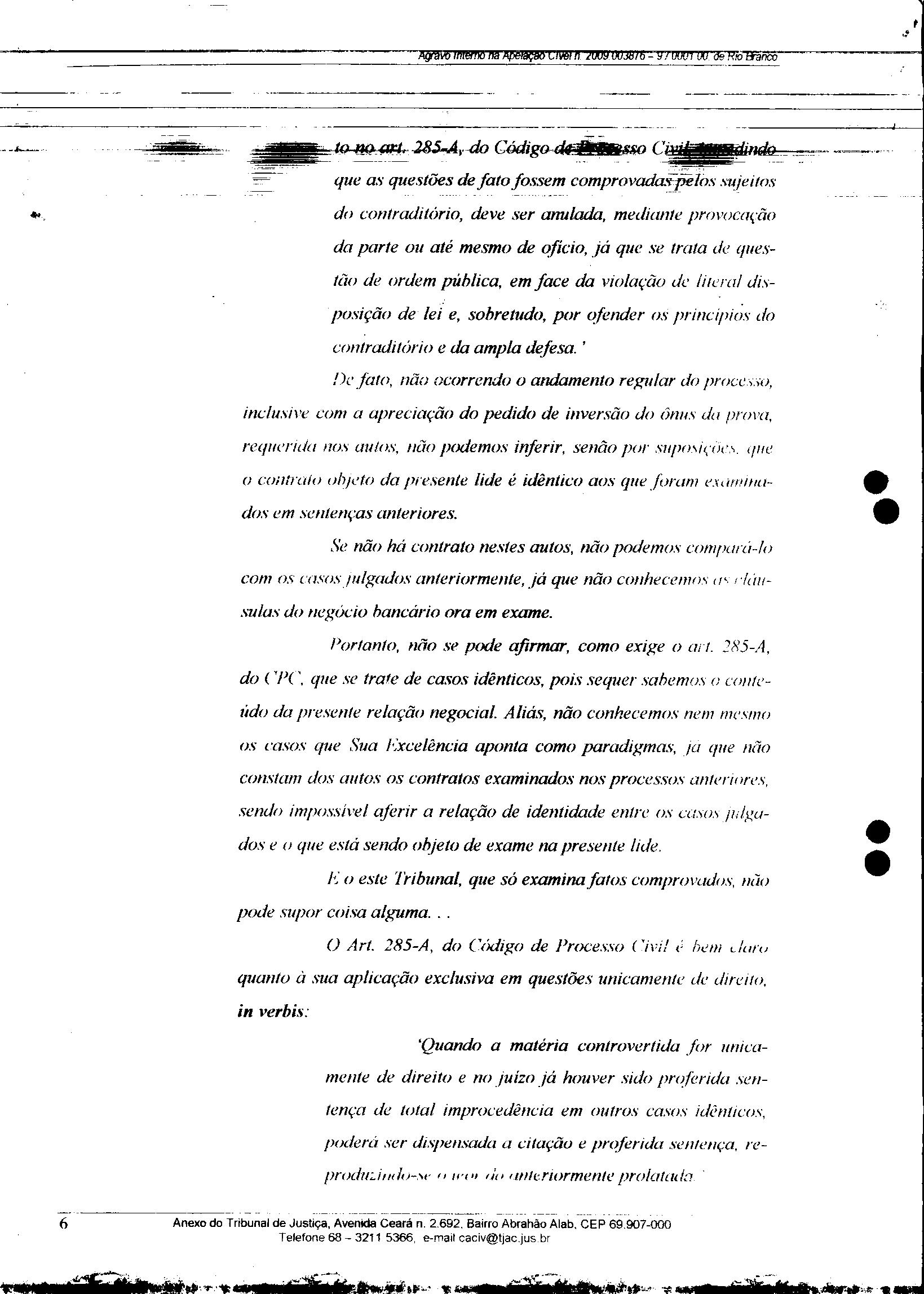}&
  \includegraphics{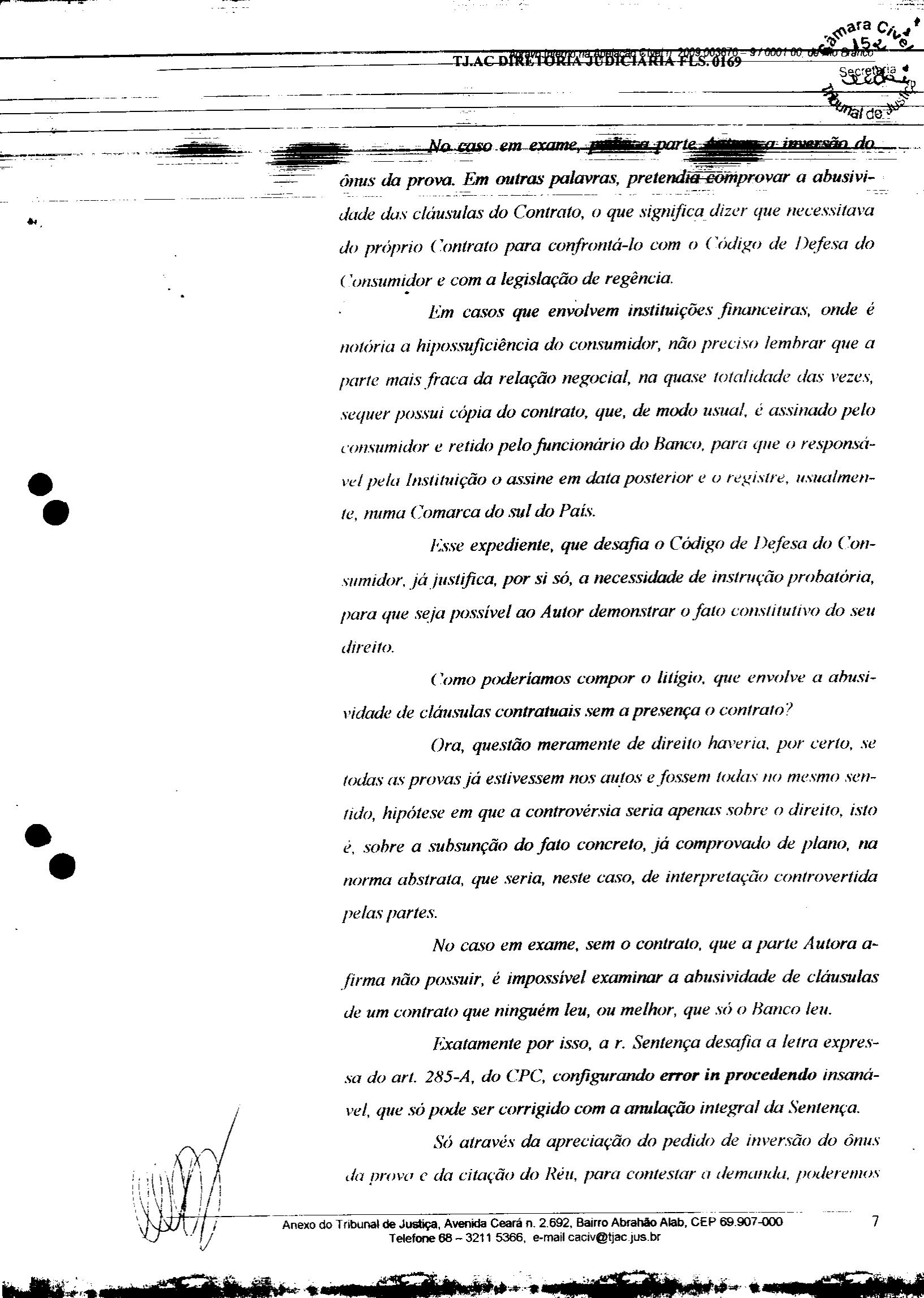}
  \end{tabularx}
  \caption{The first eight pages of a lawsuit. While the first page is clean, the others come from an older document and contain ink stains, stamps, handwritten signatures and other artifacts.}
  \label{fig:firstLSPages}
  \end{figure*}

Natural language processing (NLP) and machine learning (ML) techniques can improve this scenario by enabling faster and more efficient document analysis. Brazil's Supreme Court receives roughly 42 thousand cases each semester, which takes about 22 thousand hours for humans to sort through~\cite{stf_victor_news}. This time could be better spent on more complex stages of the workflow, such as those requiring legal reasoning. The cases reach the court as mostly unstructured and unindexed PDF files of raster-scanned documents~\cite{luzDeAraujo_etal_VICTOR_LREC_2020}. Intra-class diversity and document quality are the main challenges: the documents range from petitions and evidence to rulings and orders, originate from different Brazilian courts and often contain visual noise such as handwritten annotation, stamps, and stains (Figure~\ref{fig:firstLSPages}).

Therefore, our goal is to explore and evaluate methods that automatically classify document pages by combining different sources of information. Though previous work~\cite{luzDeAraujo_etal_VICTOR_LREC_2020,mota_etal_cnnPet_ENIAC_2020} has examined Brazilian legal document classification, we are the first to combine visual, textual, and sequential data to train better performing models. Our main contributions are:

\begin{enumerate}
  \item a multimodal dataset of lawsuits composed of ordered document images and corresponding texts.
  \item proposing and evaluating two mutimodal combination methods and two sequence learning methods that leverage textual, visual and sequential information to improve Supreme Court document classification.
  \item outperforming the state-of-the-art results on the small version of the VICTOR dataset~\cite{luzDeAraujo_etal_VICTOR_LREC_2020}.
\end{enumerate}

The rest of this paper is organized as follows. In Section~\ref{sec:revision}, we examine previous work on multimodal document classification. In Section~\ref{sec:data}, we describe the data we used to train and evaluate our models, which we describe, along with their experimental setting, in Section~\ref{sec:methods}. We then discuss the results obtained and conclude the paper in sections~\ref{sec:disc} and~\ref{sec:conc}.

\section{Related Work}
\label{sec:revision}
Textual and visual content are two of the four document aspects listed by Chen and Blostein~\cite{chen_and_blostein_diaSurvey_IJDAR_2007} as possible feature sources. Image features range from fixed descriptors such as pixel density at different locations and scales~\cite{rusinol_etal_multimodalAdmDocument_IJDAR_2014} to approaches based on convolutional neural networks
~\cite{jain_et_wigington_ICDAR_2019,engin_etal_multiModalBank_ICAIMM_2019,audebert_etal_multiModalDocs_arxiv_2019,wiedemann_heyer_multiModalPSS_LRE_2019,mota_etal_cnnPet_ENIAC_2020} 
such as VGG-16~\cite{simonyan_etal_vgg16_2015} and MobileNetV2~\cite{sandler_etal_mobileNetV2_CVPR_2018}. Text features range from traditional methods such as latent semantic analysis~\cite{lsa} to pre-trained word embeddings (e.g.\ Fasttext~\cite{bojanowski_etal_fasttext_arxiv_2016}) and deep learning approaches~\cite{engin_etal_multiModalBank_ICAIMM_2019,audebert_etal_multiModalDocs_arxiv_2019,wiedemann_heyer_multiModalPSS_LRE_2019}. 

These feature modalities may be used by themselves or combined to improve classification performance. 
This can happen at the feature level (early fusion) or at the prediction level (late fusion). Rusinol et al. \cite{rusinol_etal_multimodalAdmDocument_IJDAR_2014} experiment with both options, trying different methods to combine predicted probabilities for late fusion (summing, multiplying, taking the maximum and logistic regression). Jain and Wigington~\cite{jain_et_wigington_ICDAR_2019} compare a spatially-aware early fusion method with four alternative methods of feature combination: concatenation, addition, compact bilinear pooling and gated units. The spatially-aware fusion underperformed simple feature combination, with concatenation, addition, and bilinear approaches performing similarly.
Engin et al.~\cite{engin_etal_multiModalBank_ICAIMM_2019} explore late and early fusion for the classification of Turkish banking documents, finding that both outperform unimodal methods. Mota et al.~\cite{mota_etal_cnnPet_ENIAC_2020} investigate multimodal classification of Brazilian court documents, concluding that multimodal approaches compare favourably with unimodal ones.

Fewer works have explored incorporating sequential information. Rusinol et al.~\cite{rusinol_etal_multimodalAdmDocument_IJDAR_2014} use an $n$-gram model of the page stream that conditions page predictions on the types of the $n-1$ previous pages to capture their sequential nature. Wiedemann and Heyer~\cite{wiedemann_heyer_multiModalPSS_LRE_2019} use as a feature of the target page the encoding of its predecessor. Luz de Araujo et al.~\cite{luzDeAraujo_etal_VICTOR_LREC_2020} feed the predictions of a text classifier to a linear-chain conditional random field (CRF) to jointly predict pages of lawsuits.

In this paper, we focus on the sequentially-aware early fusion of visual and textual features for legal document classification. To the best of our knowledge, this is the first work that considers both visual and textual modalities and sequential dependencies when classifying documents from the legal domain and in Portuguese---Luz de Araujo et al.~\cite{luzDeAraujo_etal_VICTOR_LREC_2020} do not use visual features, while Mota et al.~\cite{mota_etal_cnnPet_ENIAC_2020} do not leverage sequential information.

\section{Data}
\label{sec:data}
We perform our experiments on a dataset of 6,510 Extraordinary Appeals, comprising 339,478 pages. Each instance is a lawsuit as it is received by the Brazilian Supreme Court, before it is processed and judged. Each lawsuit contains several different documents (petitions, rulings, orders) and is represented as an ordered sequence of pages containing text.

This dataset is an extension of Small Victor~\cite{luzDeAraujo_etal_VICTOR_LREC_2020}, which we expanded to include, in addition to textual data, the document images. Every page in the expanded corpus is stored in at least one of two formats. First, as text extracted through optical character recognition~\cite{smith_Tesseract_icdar2007}, with the following additional preprocessing steps: lower-casing, removal of stop words and alphanumeric tokens, e-mail and URL tokenization, and special tokenization of legislation references (e.g., Lei (law) 11.419 to LEI\_11419). Second, as JPEG images converted from the original PDF files, with mean width and height of 1664 and 2322 pixels respectively. 

The dataset was manually annotated with labels that assign to each page the document class it belongs to. There are six possible classes: 
\begin{enumerate}
  \item \emph{Acórdão}, for lower court decisions under review;
  \item \emph{Recurso Extraordinário} (RE), for appeal petitions;
  \item \emph{Agravo de Recurso Extraordinário} (ARE), for motions against the appeal petition; 
  \item \emph{Despacho}, for court orders;
  \item  \emph{Sentença} for judgements; and
  \item  \emph{Others} for documents not included in the previous classes.
\end{enumerate} 

Most of the samples contain both textual and visual sources of information, except for 33,849 images with no corresponding text and 4 texts with no corresponding image. We discuss our strategy for dealing with missing data when training fusion models in Section~\ref{subSec:methodsFusion}. The corpus is divided into train, validation and test splits containing 70\%, 15\%, and 15\% of all suits, respectively. Table~\ref{table:classDist} presents the number of text and image samples across data splits and classes. Due to the nature of the data, a document may appear more than once in a lawsuit, so we present both raw and deduplicated counts. That said, given that the corpus has been split by lawsuits, there is no sample intersection between splits.

\begin{table*}
  \centering
  \small
  \caption{Class counts per split, showing the number of page images and text extracted through OCR. Between parentheses, the deduplicated counts.}\label{table:classDist}
  \begin{tabularx}{\textwidth}{ X r r r r r r }
  \toprule
   Class &  \multicolumn{2}{c}{Training set} & \multicolumn{2}{c}{Validation set} & \multicolumn{2}{c}{Test set}\\
   \cmidrule(lr){2-3} \cmidrule(lr){4-5} \cmidrule(lr){6-7}
   & Image & Text & Image & Text & Image & Text \\
  \midrule
  \textit{Acórdão} &        583 (583) &     553 (553) &        320 (314) &    299 (293) &      287 (285) &    273 (271) \\
  ARE &       4,258 (4,220) &   2,546 (2,508) &       2,798 (2,650) &  2,149 (2,001) &      2,655 (2,537) &  1,841 (1,723) \\
  \textit{Despacho} &        361 (361) &     346 (346) &        189 (183) &    183 (177) &     199 (198) &    198 (197) \\
  Others &     144,583 (140,786) & 134,134 (130,337) &    95,602 (91,434) & 84,104 (79,936) &   92,529 (87,902) & 85,408 (80,781) \\
  RE &       10,225 (10,181) &   9,509 (9,465) &        6,987 (6,803) &  6,364 (6,180) & 6,386 (6,177) &  6,331 (6,122) \\
  \textit{Sentença} &     2,177   (2,177) &   2,129 (2,129) &        1,681 (1,613) &  1,636 (1,568) &      1,503 (1,478) &  1,475 (1,450) \\
  \bottomrule
  \end{tabularx}
  \end{table*}

\begin{figure*}
  \setkeys{Gin}{width=\linewidth}
  \begin{tabularx}{\textwidth}{XXXXXX}
    {\includegraphics{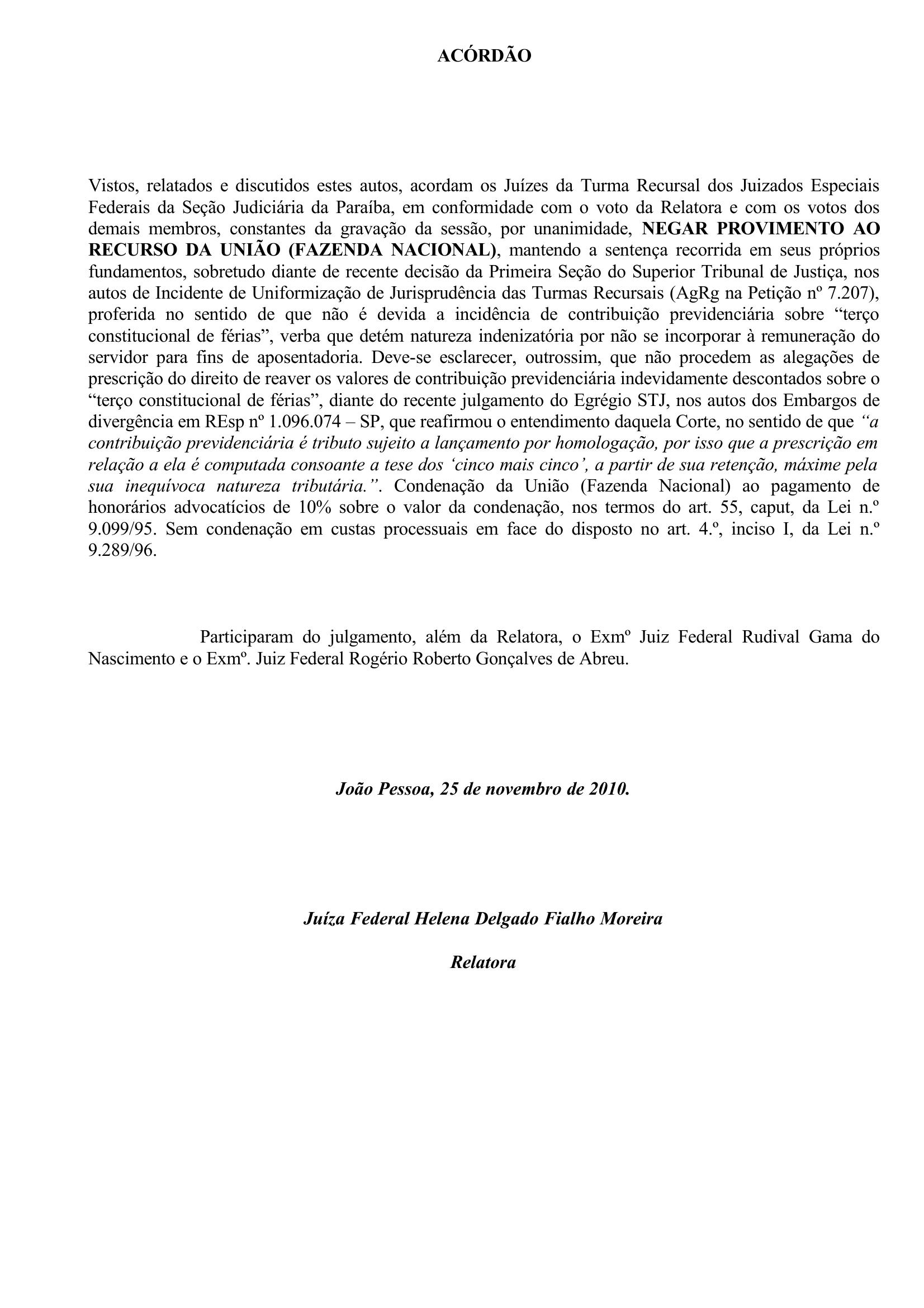}} &
    {\includegraphics{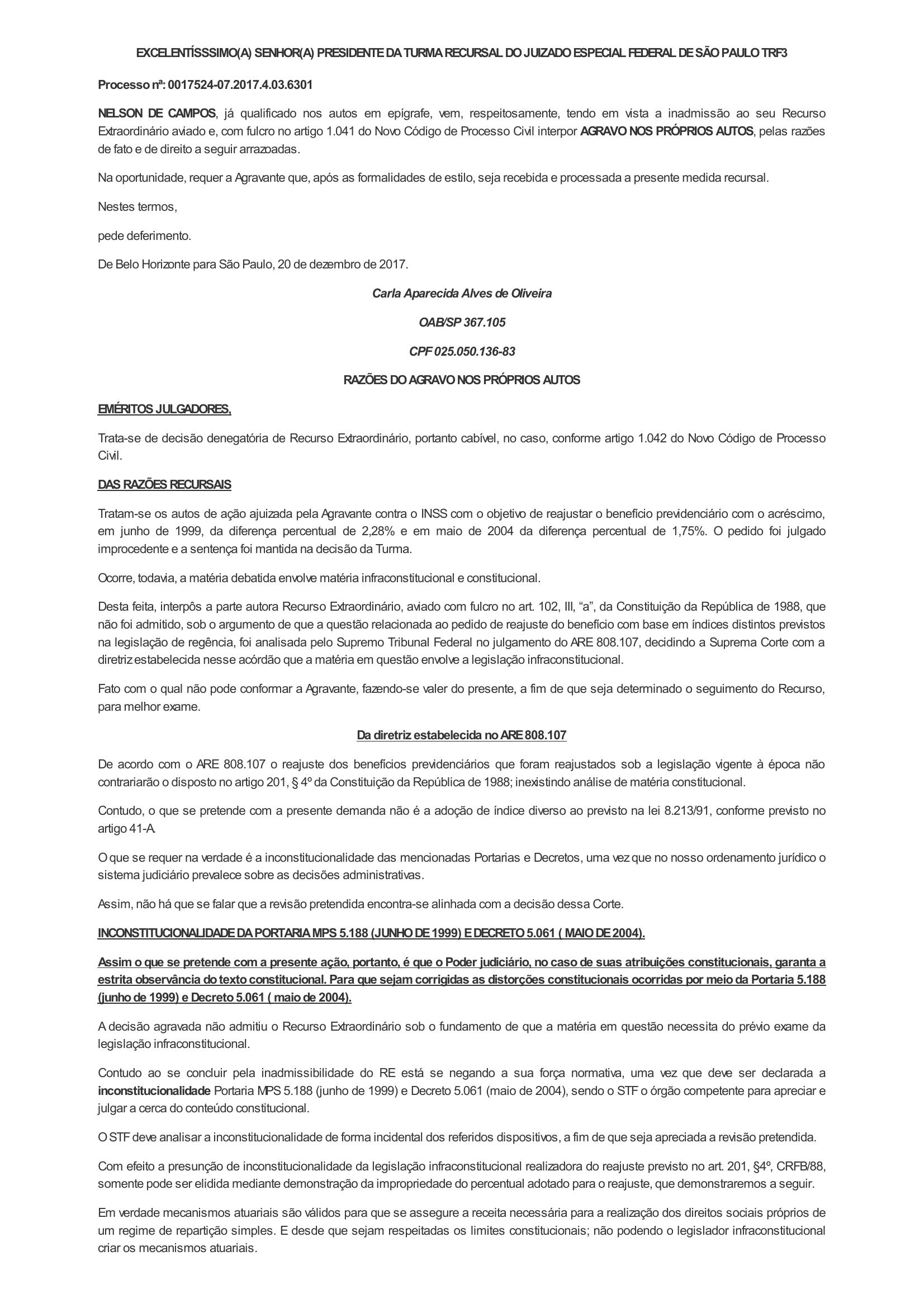}} &
    {\includegraphics{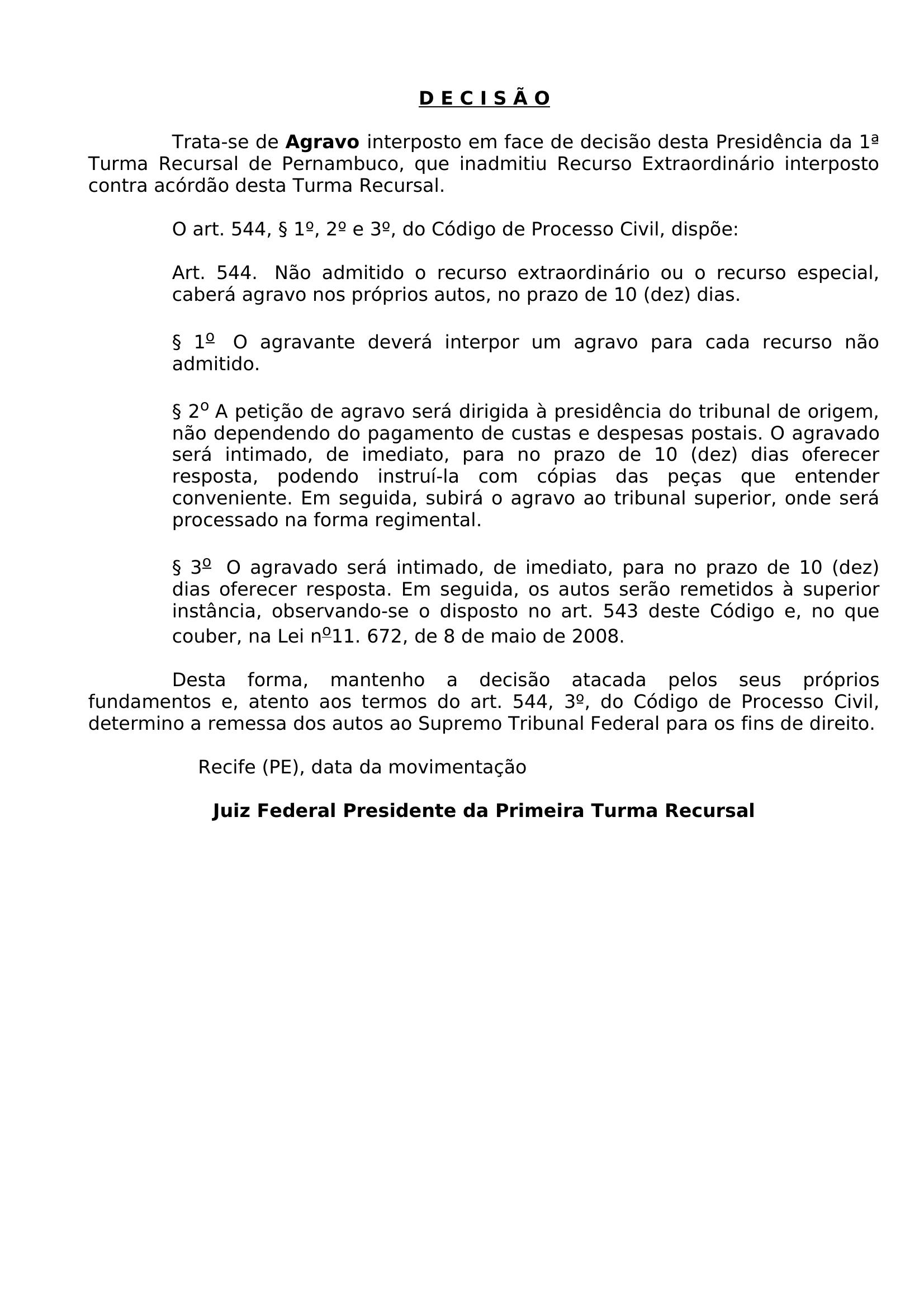}} &
    {\includegraphics{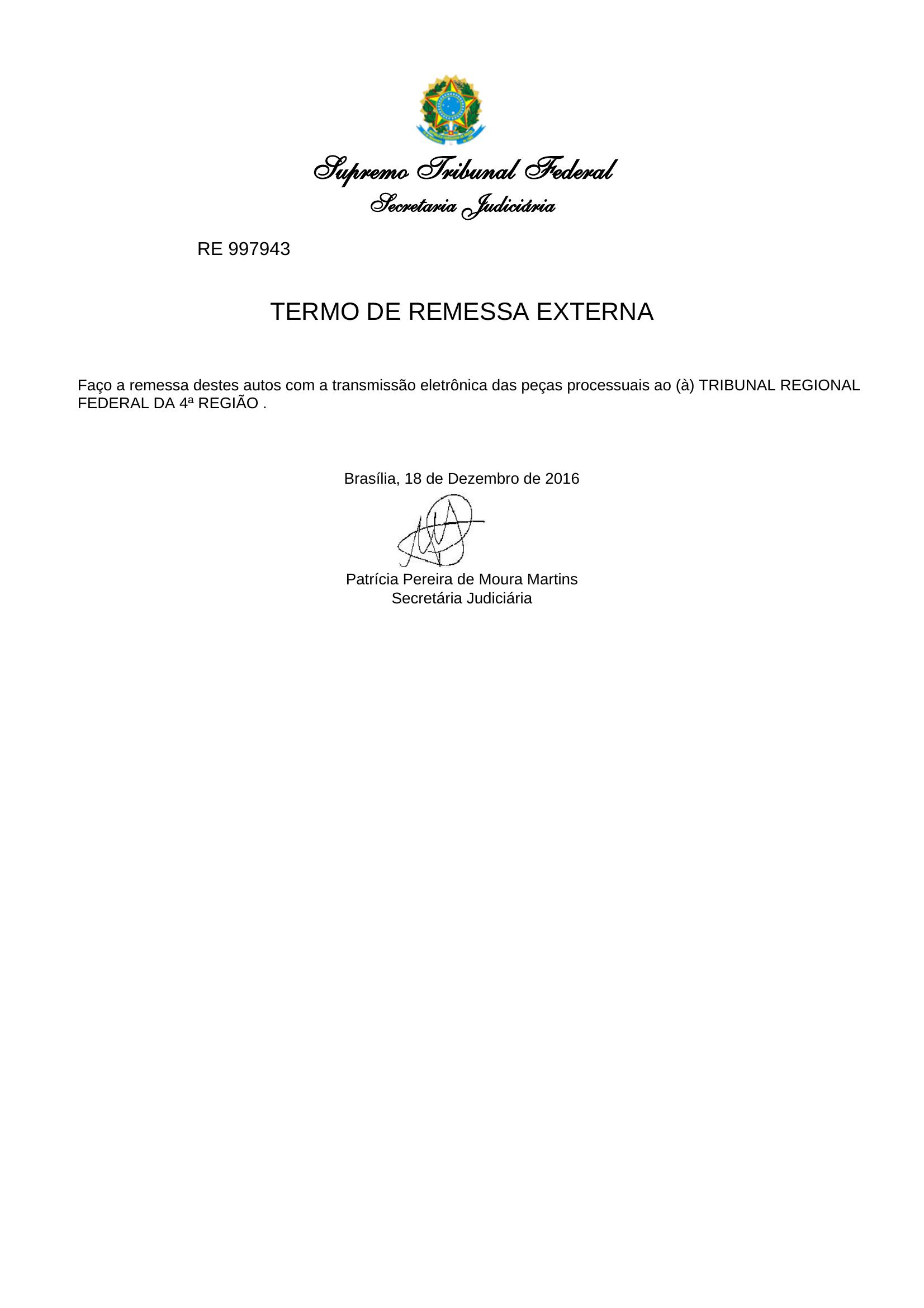}} &
    {\includegraphics{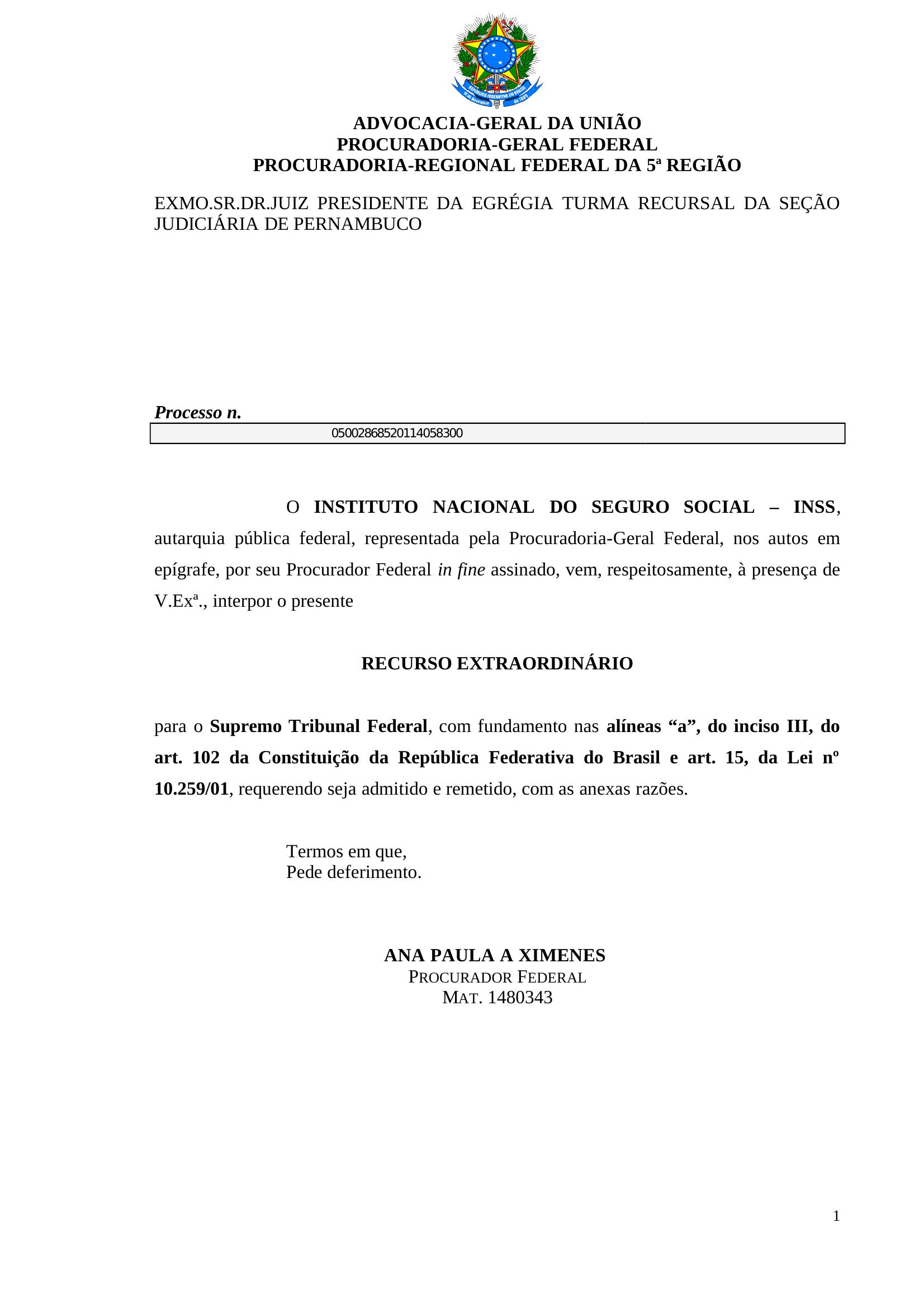}} &
    {\includegraphics{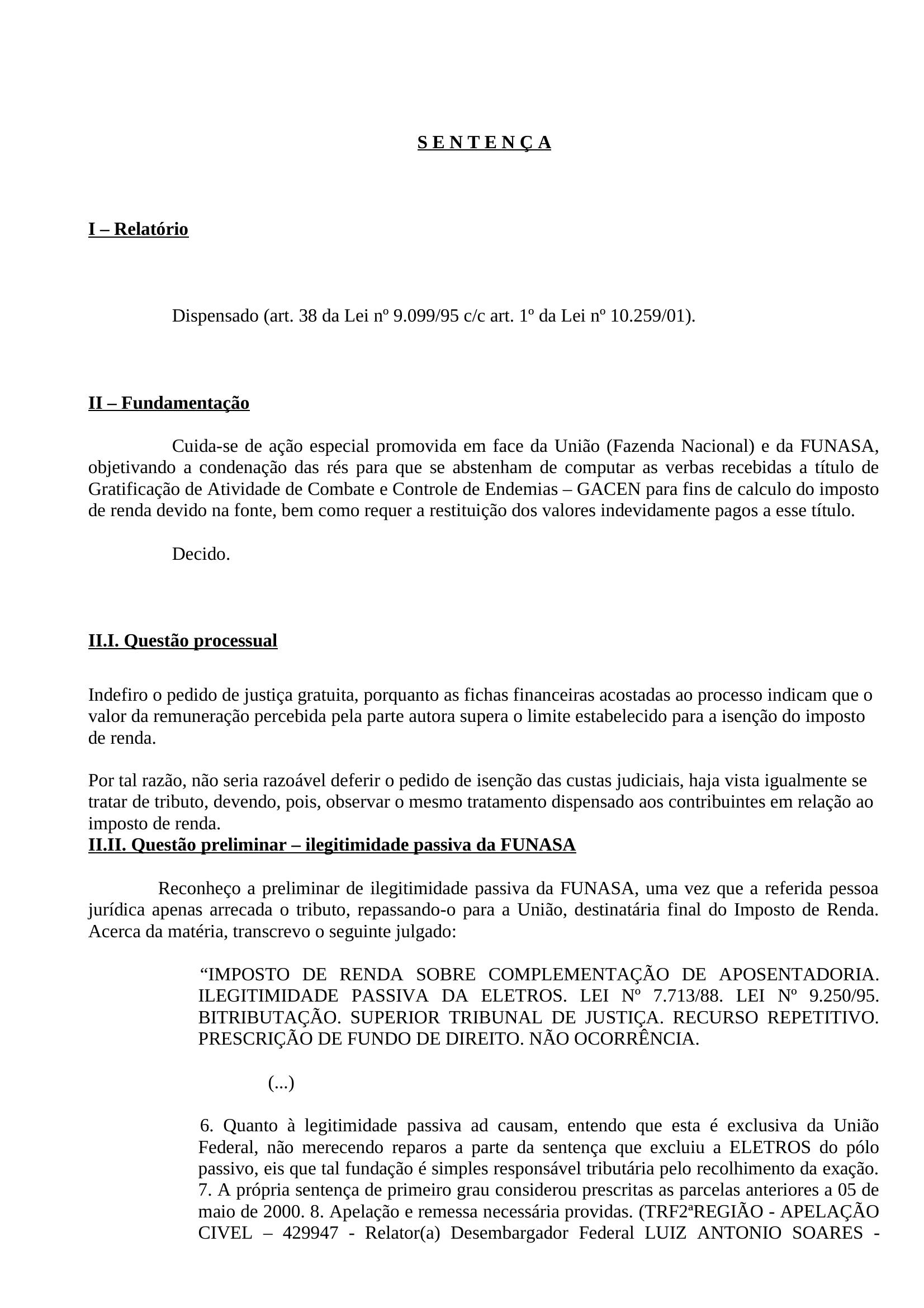}} \\

    \subfloat[\textit{Acórdão}]{\includegraphics{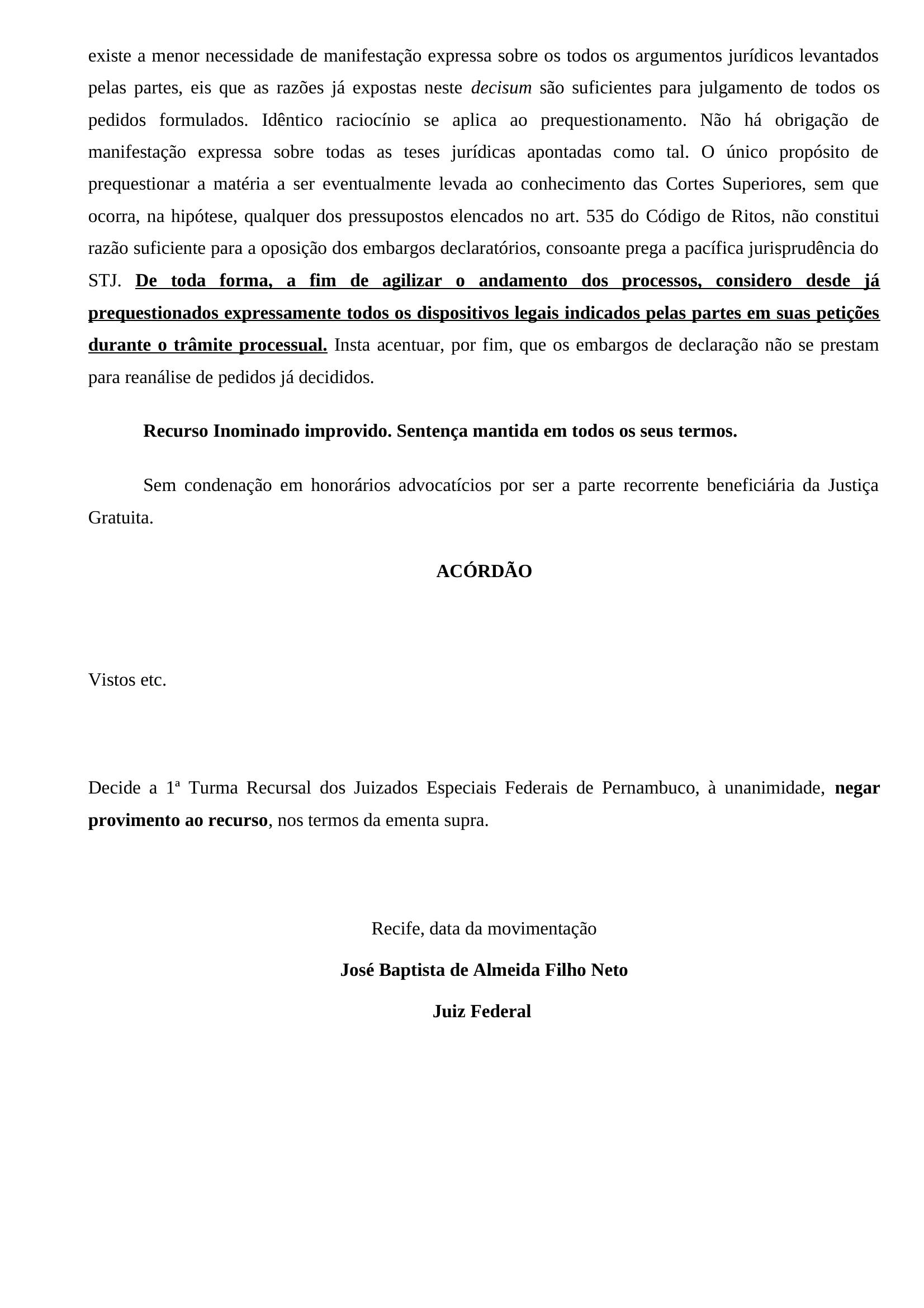}} &
    \subfloat[ARE]{\includegraphics{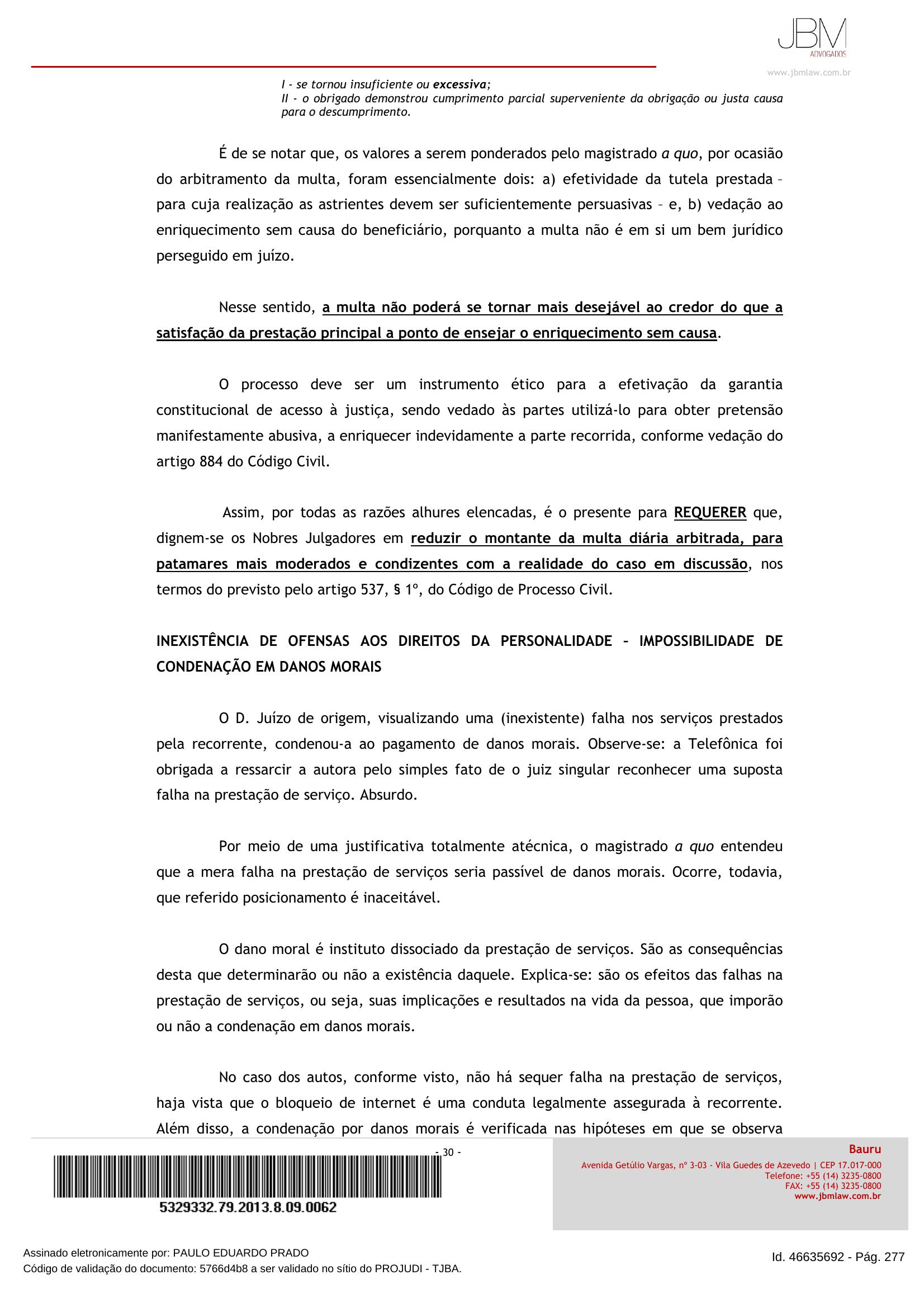}} &
    \subfloat[\textit{Despacho}]{\includegraphics{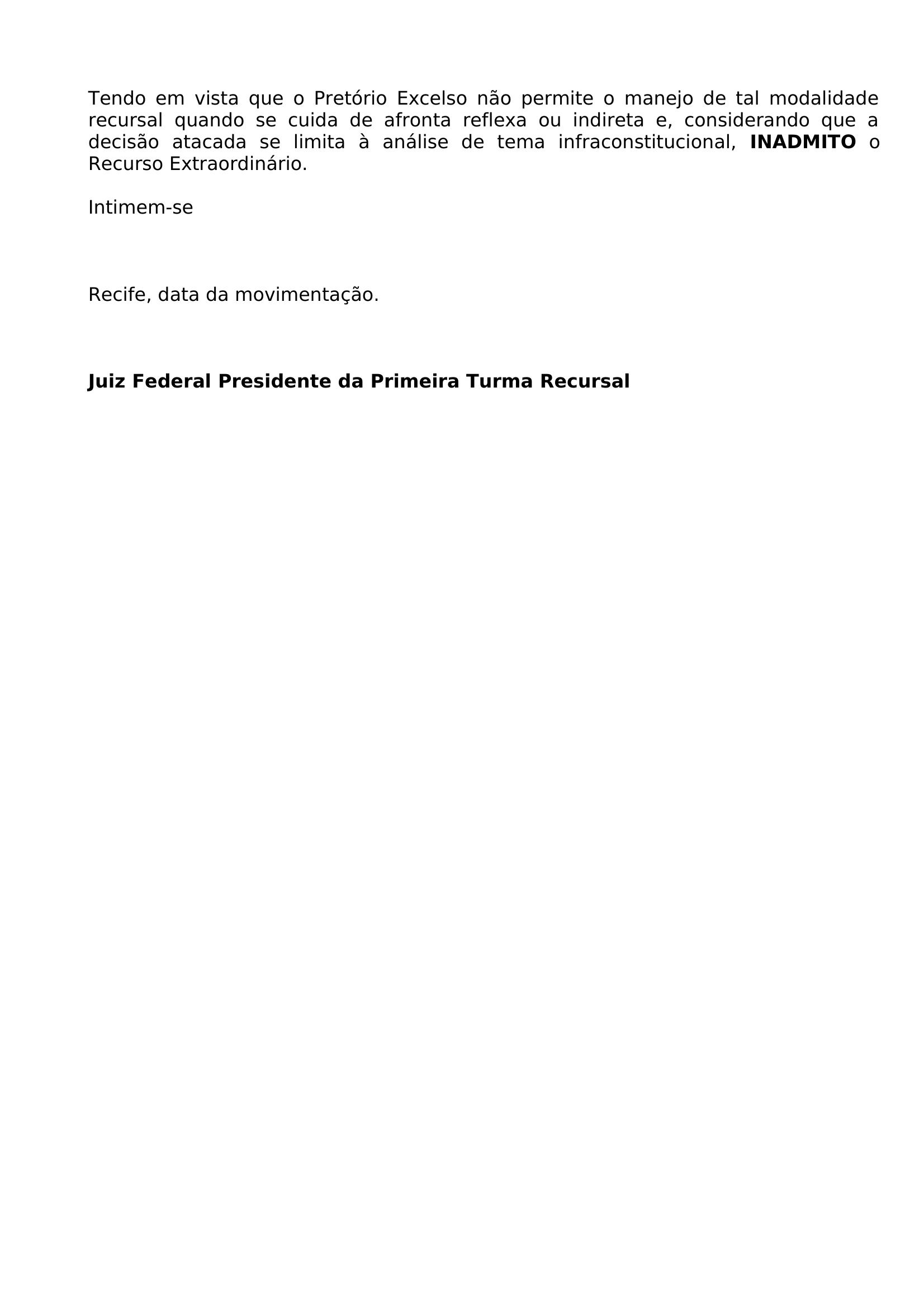}} &
    \subfloat[Others]{\includegraphics{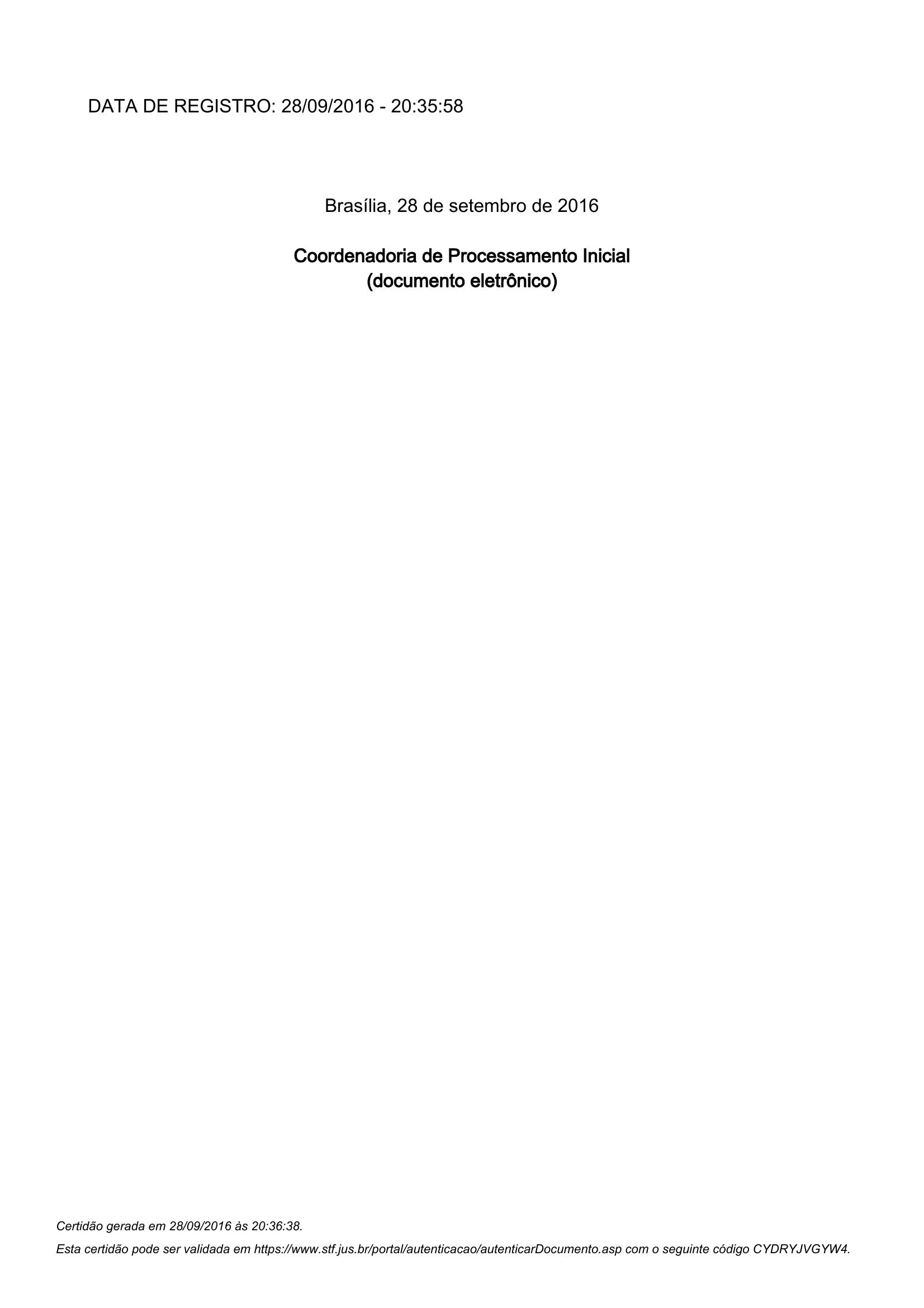}} &
    \subfloat[\textit{Petição}]{\includegraphics{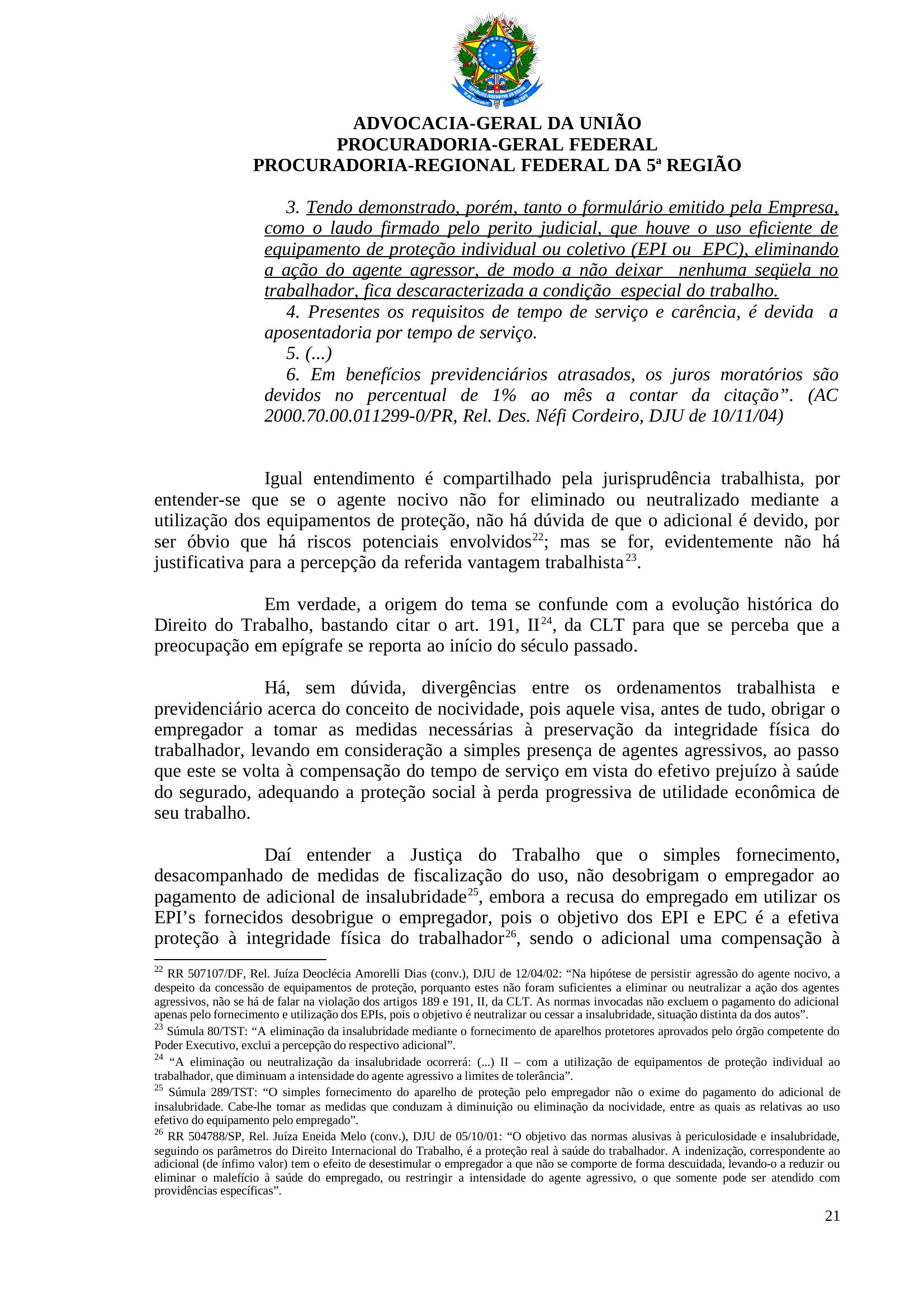}} &
    \subfloat[\textit{Sentença}]{\includegraphics{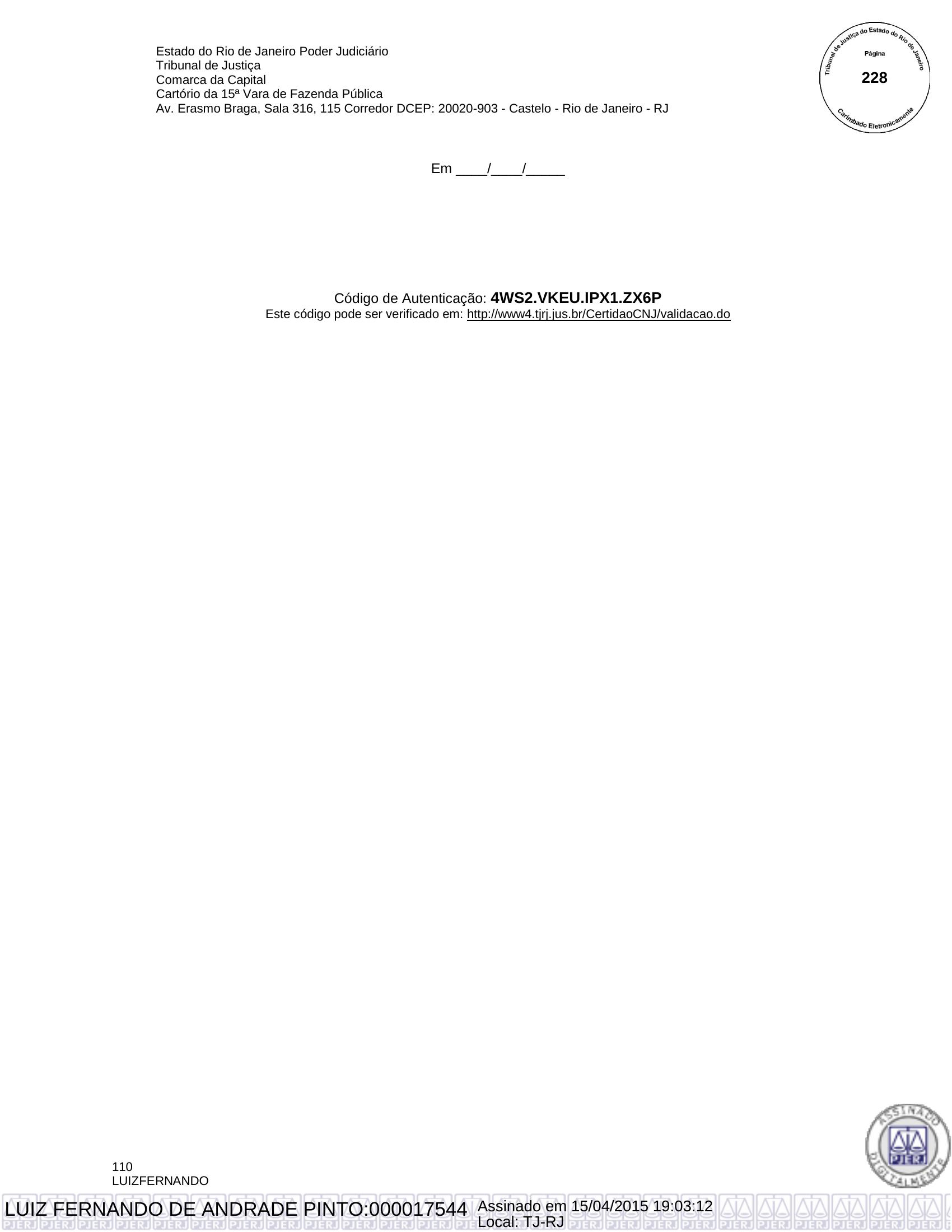}}
  \end{tabularx}
  \caption{First page (top row) and interior page samples for each class.}
  \label{fig:firstPages}
  \end{figure*}

Human agents find the first page of documents easier to classify when compared to interior pages. This is true considering both visual and textual aspects, since first pages contain highly informative cues, such as headers and titles. Figure~\ref{fig:firstPages} compares first page and interior page samples for each class. We validate this intuition in Section~\ref{subSec:firstPage}.

\section{Methods}
\label{sec:methods}
In this section we describe our methods for visual and textual page classification, feature fusion and sequence learning. We also describe the corresponding experimental settings. Unless stated otherwise, we optimise the cross-entropy loss using Adam~\cite{kingma2014adam} and mini-batches of 64 samples. As metrics, we report arithmetic (average) and weighted by class frequencies (weighted) means of the \fone~score, defined as:
\begin{equation}
  \fone = 2\cdot\frac{\text{precision} \cdot \text{recall}}{\text{precision} + \text{recall}}\,,
\end{equation}
whereas precision and recall are defined as follows: let tp, fp and fn be the number of true positives, false positives and false negatives, respectively. Then,
\begin{align}
    \text{precision}&=\frac{\text{tp}}{\text{tp} + \text{fp}},\\
    \text{recall}&=\frac{\text{tp}}{\text{tp} + \text{fn}}\,.
\end{align}
 We evaluate the models using the parameters with the best average \fone~score computed on the validation set. That is, after each epoch, we only save model parameters if the validation performance is the highest up to that point.

\subsection{Text Classification}
\label{subSec:methodsText}
\begin{figure}[htbp]
  \includegraphics[width=\linewidth]{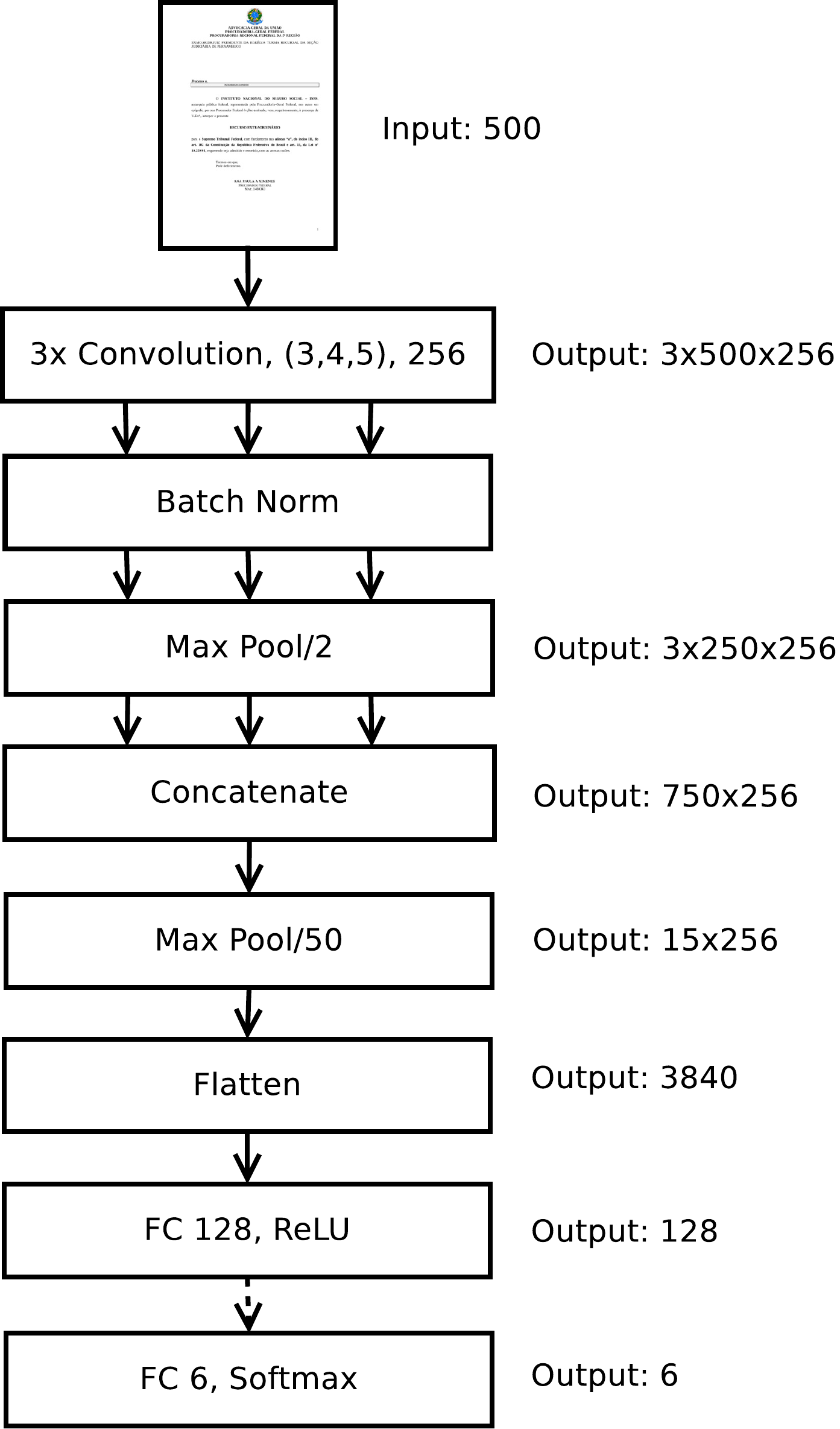}
  \caption{The network architecture we used for text classification. Dashed lines indicate dropout was applied.}
  \label{fig:cnn}
  \end{figure}

We use the CNN architecture described by Luz de Araujo et al.~\cite{luzDeAraujo_etal_VICTOR_LREC_2020} as the method for text classification. It is a shallower version of the one proposed by Conneau et al.~\cite{veryDeepTextClf} and works on the word level instead of the character level. Figure~\ref{fig:cnn} summarises the CNN structure. To process a sample, the network takes as input its first 500 tokens and embeds them into 100-dimensional vectors. The vectors are then fed to three convolutional blocks composed of a convolutional layer with three sets of 256 one-dimensional filters with different sizes (3, 4 and 5) followed by batch normalisation and max-pooling of size 2. The resulting vectors are concatenated and fed to a max pool layer of size 50. The output is flattened and processed by two fully connected (FC) layers, with the softmax function producing the prediction. Dropout is applied to the output of the first FC layer with a dropping probability of 50\%.

As a strategy to deal with class imbalance, we train a variant of the CNN model, which we call CNN-w, that weighs each sample contribution to the loss by a factor inversely proportional to its class frequency. Let $c$ be the number of classes and $\mathbf{w}$ a $c$-dimensional vector whose component $i$ is the factor for class $i$. Then the factors are computed by the following equation, as implemented in the scikit-learn library~\cite{buitinck_etal_sklearn_ECML_2013}:

\begin{equation}
  w_i=\frac{n}{c \cdot f_i}\,,
\end{equation}
where $n$ is the number of training samples and $f_i$ is the number of samples from class $i$.

\subsection{Image Classification}
\label{subSec:methodsImage}
To classify document images, we fine-tune a ResNet50~\cite{he_etal_resnet_CVPR_2016} model pretrained on ImageNet~\cite{russakovsky_etal_imagenet_IJCV_2015}. We first train only the head of the model for one epoch, employing a cyclic learning rate with cosine annealing~\cite{cyclelr}.
Then, we train all layers for one cycle of 6 epochs with discriminative fine-tuning~\cite{ulmfit}. As we did for the text classifier, we train a variant of the model with factors inversely proportional to class frequency: ResNet50-w.

To choose learning rates, we use the learning rate range test \cite{lrfinder}. That is, we train the model for a few iterations, starting from a low learning rate value and increasing it after each mini-batch, plotting the loss against the learning rates. We then pick a learning rate close to the point where the loss starts to increase---high enough for quick learning, but not so high as to impede learning.

\subsection{Image and Text Combination Strategies}
\label{subSec:methodsFusion}
In this section we describe our proposed method for early fusion of visual and textual features, a baseline method for comparison and an ablation analysis of the fusion classifier.

\subsubsection{Hybrid Classifier}
\begin{figure}
  \includegraphics[width=\linewidth]{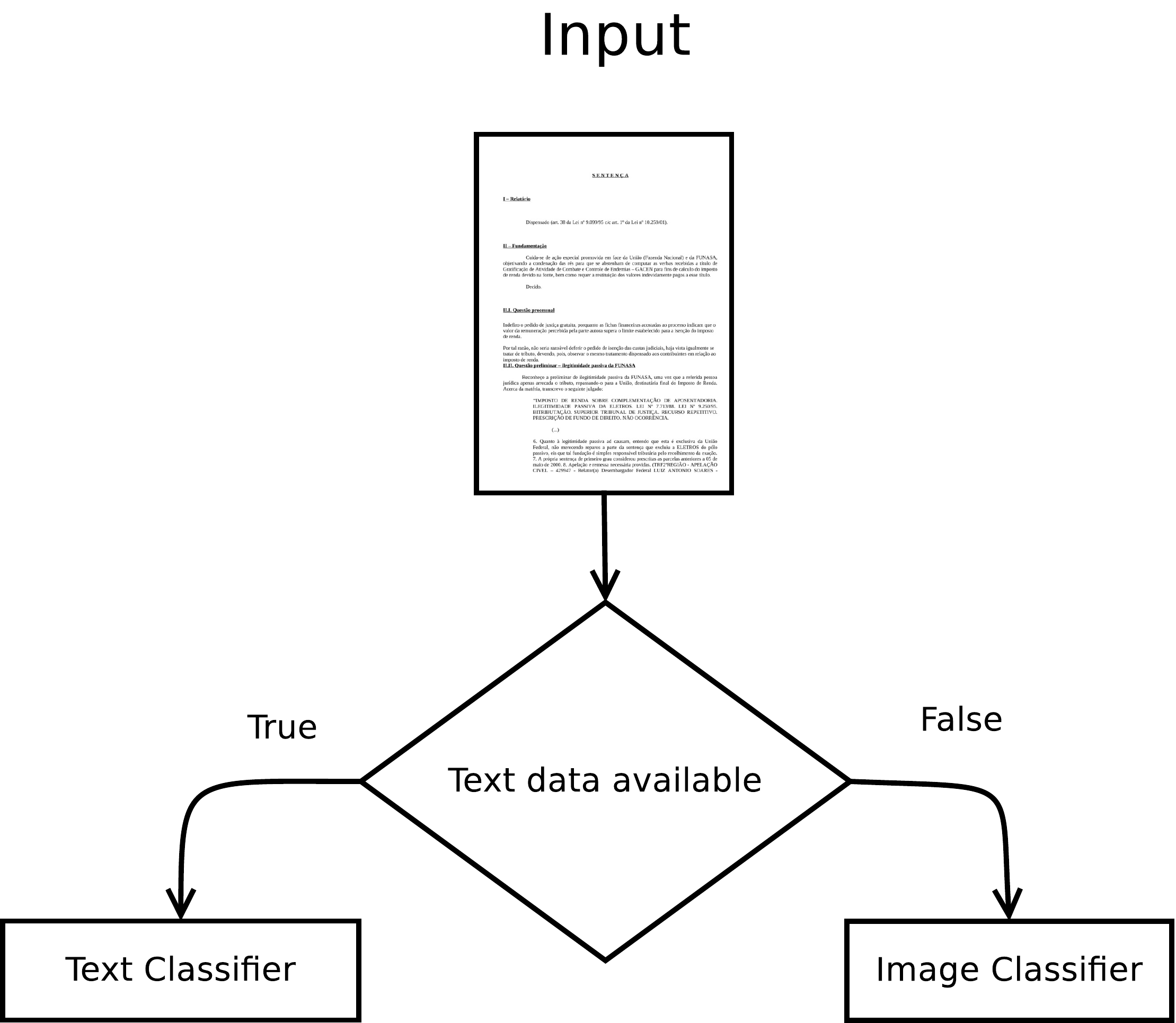}
  \caption{The hybrid classifier (HC): a baseline fusion classifier that only uses visual information if text data is not available.}
  \label{fig:HC}
  \end{figure}

As a baseline method that fuses visual and textual data we use a hybrid classifier (HC) that works as follows: if textual data is available, use the best text classifier; otherwise, use the best image classifier. The intuition is that this approach would be at least as good as using only text data, which better discriminates the document classes when compared with visual data. Figure~\ref{fig:HC} illustrates the method.

\subsubsection{Fusion Module}
To fuse visual and textual data we first compute representations using the trained text and image classifiers. As text embeddings, we extract the 3,840-dimensional activations of the flatten layer of the CNN (Fig.~\ref{fig:cnn}). As image embeddings, we take the activations of the last convolutional block in the ResNet and apply global average and global max pooling. Then we concatenate and flatten the result, obtaining 4,096-dimensional vectors.

\begin{figure}
  \includegraphics[width=\linewidth]{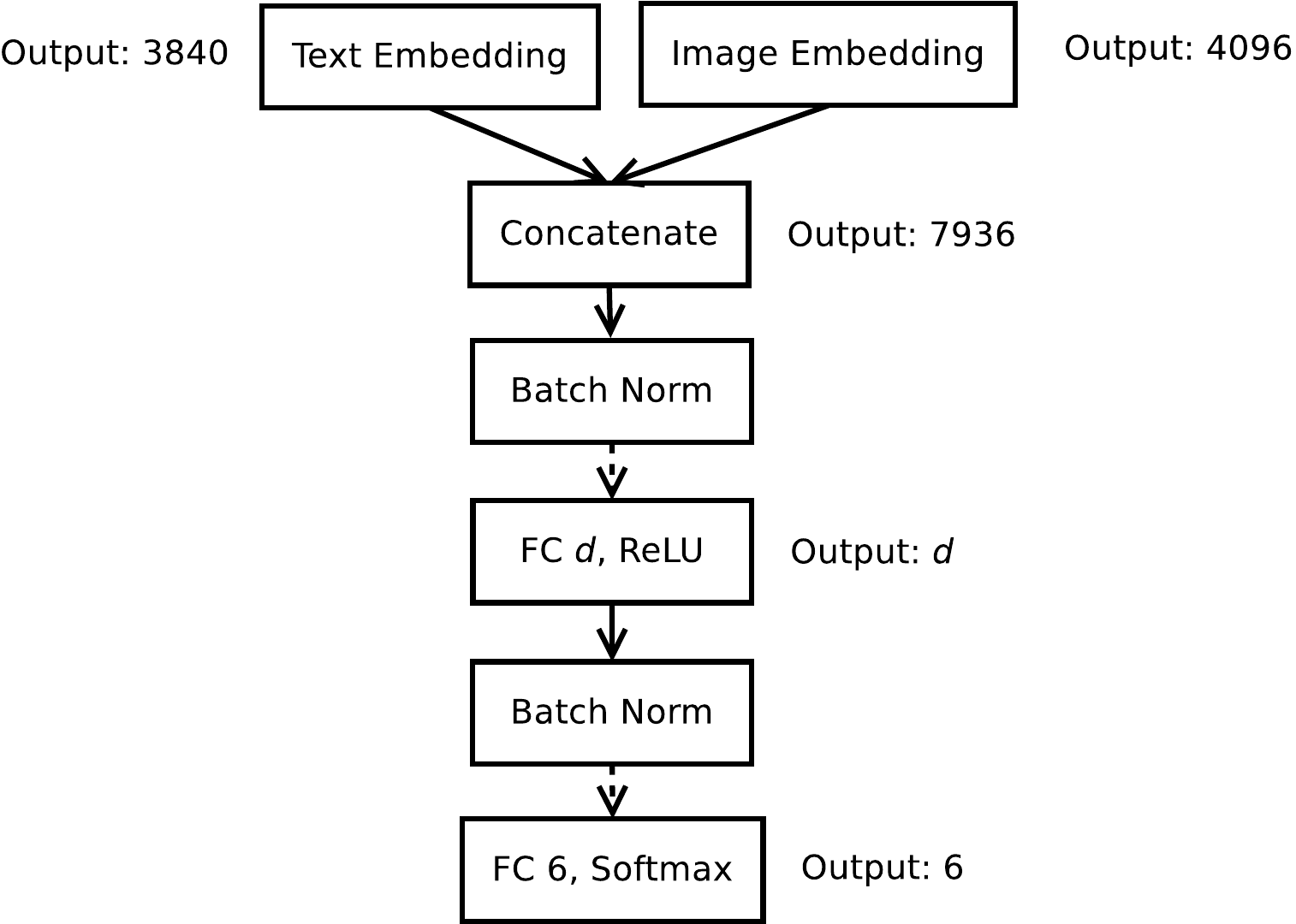}
  \caption{Fusion Module (FM): the proposed method for early fusion of textual and visual information. Dashed lines indicate dropout was applied. The hyperparameter $d$ is the number of units in the first fully connected layer.}
  \label{fig:fm}
  \end{figure}

The pre-computed representations are concatenated and fed to a batch normalisation layer~\cite{batchnorm}, followed by an FC layer with $d$ units, batch normalisation, and a final FC layer. The softmax function produces the predictions. Figure~\ref{fig:fm} illustrates our proposed Fusion Module (FM).

In case of missing data, when only the document image or text is available---but not both---we experiment with two options. The first uses learnable embeddings for missing text or image; the other, simply uses a vector of zeroes in such cases (FM-zero).

We run preliminary experiments with one cycle of 10 epochs for each of four configurations, varying the number of hidden units $d$ (128 or 512) and the use of learnable embeddings or zero vectors for missing data. We then train the model that obtained the highest average \fone~score from scratch for one cycle of 20 epochs. Learning rates are chosen using the range test.

\subsection{Sequence Classification}
\label{subSec:methodsSequence}
Given that a lawsuit is composed of an ordered series of pages, one can, instead of classifying each page by itself, leverage the sequential nature of the data by treating the problem as a sequence labelling task. That is, rather than having a page and a class prediction as input and output, the sequence classification approach outputs a sequence of class predictions, given a sequence of input pages. We employ the IOB tagging scheme~\cite{IOB_scheme} to better leverage the sequential information: we prepend ``B-'' to the ground truth of first-page samples and ``I-'' otherwise. For example, if a suit begins with a RE of three pages followed by an ARE of equal length, the label sequence would start with B-RE, I-RE, I-RE, B-ARE, I-ARE, I-ARE.

\subsubsection{CRF postprocessing}
\begin{figure}
  \includegraphics[width=0.8\linewidth]{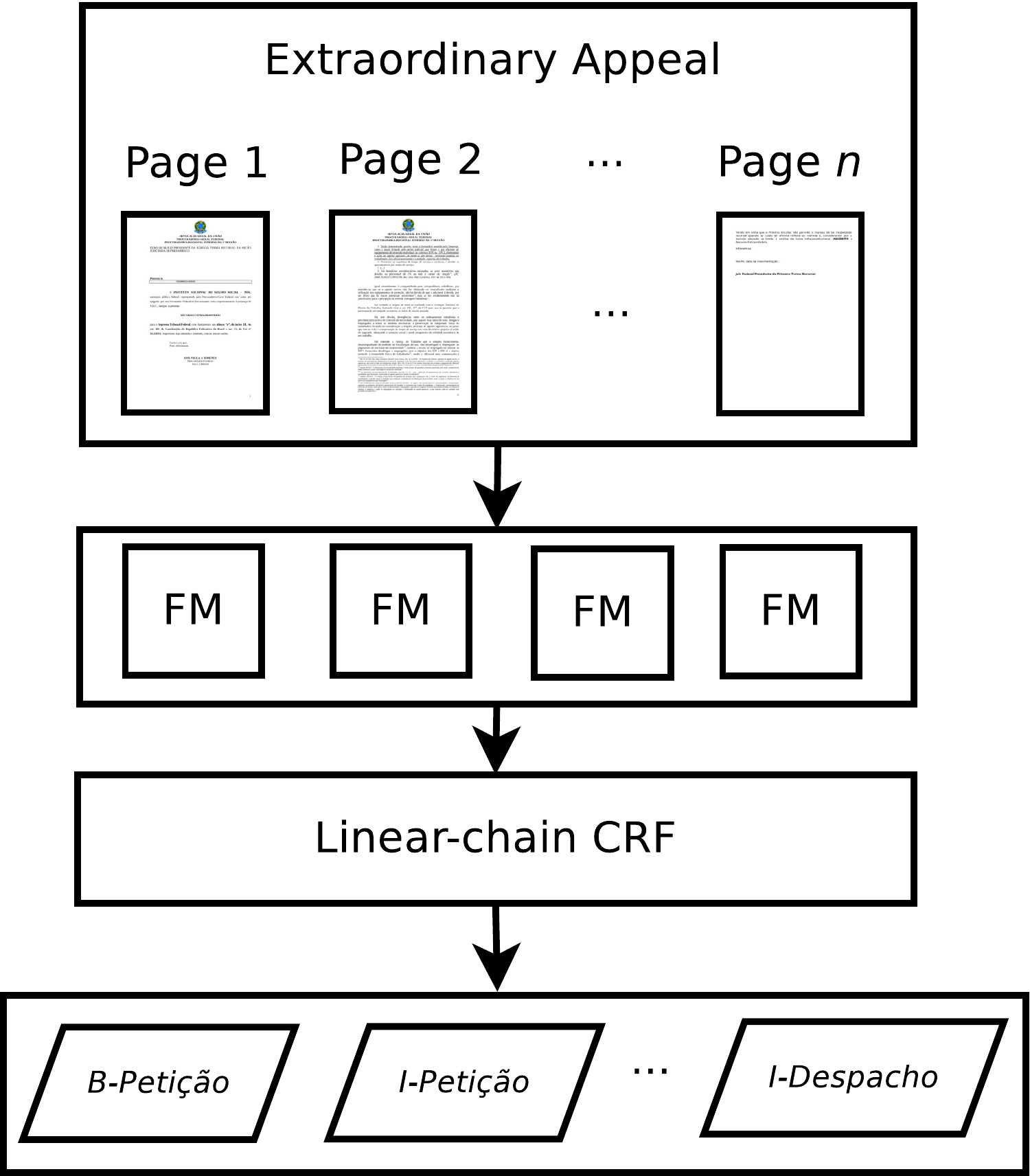}
  \caption{Baseline sequence classification method (FM+CRF). We feed the (pre-computed) predictions of the Fusion Module (FM) to a linear-chain CRF to jointly predict the class of each page in an Extraordinary Appeal.}
  \label{cnnCRF}
  \end{figure}

As a baseline method for sequence classification, we save the predictions of the FM for all samples in our data. Then we use these six-dimensional vectors as features to train a linear-chain conditional random field (CRF)~\cite{CRF}. Figure~\ref{cnnCRF} illustrates this method (FM+CRF).

  \begin{table*}[ht]
    \centering
    \small
    \caption{Test set \fone~scores (in \%) of our main approaches for image, text, fusion and sequence classification. Image results are reported for the image test set; all the others, for the text test set.}\label{table:finalResults}
    \begin{tabularx}{\linewidth}{ X r r r r r r r r}
    \toprule
      Class &  \multicolumn{2}{c}{Majority Baseline} & Text & \multicolumn{2}{c}{Image} & Fusion & \multicolumn{2}{c}{Sequence}\\
      \cmidrule(lr){2-3}\cmidrule(lr){4-4}\cmidrule(lr){5-6}\cmidrule(lr){7-7}\cmidrule(lr){8-9}
      &  Text & Image & CNN & ResNet50-w & ResNet50 & FM & FM+CRF &BiLSTM-F\\
    \midrule
    \textit{Acórdão}  & 00.00 & 00.00 &  89.96 & 18.45 & 06.78 & 90.74 & \textbf{91.56} & 88.97\\
    ARE               & 00.00 & 00.00 & 55.72 & 11.33 & 00.00 & 57.92 & 60.74 & \textbf{61.16}\\
    \textit{Despacho} & 00.00 & 00.00 & 62.94 & 08.44 & 00.00 & 63.98 & 62.69 & \textbf{64.07}\\
    Others        & 94.41 & 94.28    &  97.31 & 61.72 & 95.02 & 97.24 & \textbf{97.67} & 97.46\\
    RE                & 00.00 & 00.00 & 75.59 & 32.59 & 34.96 & 75.47 & 78.43 & \textbf{79.67}\\
    \textit{Sentença} & 00.00 & 00.00 & 80.53 & 43.52 & 48.67 & 82.04 & 83.42 & \textbf{85.26}\\
    \midrule
    Average     & 15.73 & 15.71     &  77.01 & 29.34 & 30.91 & 77.90 & 79.09 & \textbf{79.43}\\
    Weighted     & 84.41 & 84.07     &  94.72 & 58.09 & 87.67 & 94.72 & \textbf{95.38} & 95.30\\
    \bottomrule
    \end{tabularx}
    \end{table*}

    \begin{table}
      \centering
      \small
      \caption{Text classification: comparison between average and weighted by class frequencies  validation set \fone~scores (in \%) of the different approaches. The suffix \textit{-w} indicates the use of class frequency penalty weights.}\label{table:textComp}
      \begin{tabularx}{\linewidth}{ X r r}
      \toprule
       Method &  Average \fone & Weighted \fone \\
      \midrule
      CNN-w & 64.24 & 90.98 \\
      CNN & \textbf{77.14} & \textbf{94.37} \\
      \bottomrule
      \end{tabularx}
      \end{table}
    
      \begin{table}[htbp]
        \centering
        \small
        \caption{Image classification: comparison between validation set \fone~scores (in \%) of the different approaches. The suffix \textit{-w} indicates the use of class frequency penalty weights.}\label{table:imgComp}
        \begin{tabularx}{\linewidth}{ X r r}
          \toprule
          Class &  ResNet50-w & ResNet50\\
         \midrule
         \textit{Acórdão}  & \textbf{17.68} & 02.50          \\
         ARE               & \textbf{11.56} & 00.00          \\
         \textit{Despacho} & \textbf{07.53} & 00.00          \\
         Others            & 63.01          & \textbf{94.77} \\
         RE                & 32.48          & \textbf{33.03} \\
         \textit{Sentença} & 43.46          & \textbf{49.20} \\
         \midrule
         Average           & 29.29          & \textbf{29.92} \\
         Weighted          & 59.13          & \textbf{87.09} \\
        \bottomrule
        \end{tabularx}
        \end{table}
    
        \begin{table}
          \centering
          \small
          \caption{Fusion Module: impact of number of hidden units and learnable embeddings for missing data on average validation set \fone~scores (in \%). The suffix \textit{-zero} indicates the use of vector of zeros for missing data (as opposed to using learnable embeddings).}\label{table:FMcomp}
          \begin{tabularx}{\linewidth}{ X r}
          \toprule
           Method &  Average \fone\\
          \midrule
          FM-512 & 74.49 \\
          FM-512-zero & 68.02 \\
          FM-128 & \textbf{75.70} \\
          FM-128-zero & 72.95 \\
          \bottomrule
          \end{tabularx}
          \end{table}

          \begin{table*}
            \centering
            \small
            \caption{Fusion Module ablation, comparing the test set \fone~scores (in \%) of the hybrid classifier and of a version of the fusion module that ignores image activations (w/o img acts), that is, always uses the missing image embedding. For the hybrid classifier, we report results using both image classifiers: with (HC-w) and without (HC) class frequency penalty. Between parentheses, the difference in performance compared with using the original fusion module (FM).}\label{table:zeroImgAct}
            \begin{tabularx}{\linewidth}{ X r r r r r}
            \toprule
             Class & \multicolumn{2}{c}{Text test split} & \multicolumn{3}{c}{Text + image test split} \\
             \cmidrule(lr){2-3} \cmidrule(lr){4-6}
             & FM & fusion w/o img acts & FM & HC-w & HC \\
            \midrule
            \textit{Acórdão} & 90.74 & 88.27 (-2.47) & 88.5 &  41.36 (-47.14) & 87.68 (-0.82)\\
            ARE             & 57.92  & 54.09 (-3.83) & 56.6 & 49.02 (-7.58)  & 43.91 (-12.69) \\
            \textit{Despacho}& 63.98 & 62.01 (-1.97) & 63.79 & 42.71 (-21.08) & 61.85 (-1.94)  \\
            Others          & 97.24  & 97.27 (+0.03) & 97.03 & 95.80 (-1.23)  & 97.02 (-0.01) \\
            RE             & 75.47   & 73.26 (-2.21) & 75.05 & 72.11 (-2.94)  & 75.00 (-0.05) \\
            \textit{Sentença} & 82.04& 79.58 (-2.46) & 81.21 & 74.07 (-7.14)  & 79.68 (-1.53) \\
            \midrule
            Average        & 77.90   & 75.74 (-2.16) & 77.03 &  62.51 (-14.52) & 74.19 (-2.84)\\
            Weighted       & 94.72   & 94.47 (-0.25) & 94.32 & 92.58 (-1.74)  & 93.95 (-0.37) \\
            \bottomrule
            \end{tabularx}
            \end{table*}

            \begin{table}[htbp]
              \centering
              \small
              \caption{Sequence classification: comparison between average and weighted by class frequencies  validation set \fone~scores (in \%) of the different approaches. }\label{table:seqComp}
              \begin{tabularx}{\linewidth}{ X r r}
              \toprule
               Method &  Average \fone & Weighted \fone \\
              \midrule
              BiLSTM & 77.16 & 94.25\\
              BiLSTM-CRF & 78.45 & 94.46\\
              BiLSTM-F & \textbf{79.03} & \textbf{94.81} \\
              BiLSTM-F-CRF & 78.87 & 94.58 \\ 
              \bottomrule
              \end{tabularx}
              \end{table}

              \begin{table*}
                \centering
                \small
                \caption{Comparison of first page and not first page of a document classification performance. We report test set \fone~scores (in \%) for image, text, and fusion classification using as models the CNN, the ResNet50-w and the FM, respectively. Between parentheses, the number of samples.}\label{table:pageEval}
                \begin{tabularx}{\linewidth}{ X r r r r r r}
                \toprule
                 Class &  \multicolumn{2}{c}{Text} &  \multicolumn{2}{c}{Image} &  \multicolumn{2}{c}{Fusion}\\
                 \cmidrule(lr){2-3} \cmidrule(lr){4-5} \cmidrule(lr){6-7}
                 & First page & Not first page & First page & Not first page & First page & Not first page \\
                \midrule
                \textit{Acórdão}  & \textbf{92.47 (199)}    & 83.66 (74)             & \textbf{34.28 (197)}    & 08.24 (88)             & \textbf{93.40 (199)}     & 77.19 (88)     \\
                ARE               & 47.65 (213)             & \textbf{56.74 (1,628)} & 06.71 (203)             & \textbf{12.10 (2,334)} & \textbf{59.95 (213)}     & 56.28 (2,442)  \\
                \textit{Despacho} & \textbf{71.54 (147)}    & 40.43 (51)             & \textbf{12.59 (146)}    & 03.68 (52)             & \textbf{71.81 (147)}     & 40.45 (52)     \\
                Others            & \textbf{99.02 (25,744)} & 96.58 (59,664)         & \textbf{78.19 (24,193)} & 54.29 (63,709)         & \textbf{99.04 (25,744)}  & 96.26 (66,789) \\
                RE                & 74.45 (312)             & \textbf{75.65 (6,019)} & 18.28 (301)             & \textbf{33.72 (5,876)} & \textbf{75.50 (312)}     & 75.03 (6,074)  \\
                \textit{Sentença} & \textbf{81.47 (265)}    & 80.32 (1,210)          & 26.61 (262)             & \textbf{49.71 (1,216)} & \textbf{83.11 (265)}     & 80.78 (1,238)  \\
                \midrule
                Average           & \textbf{77.77 (26,880)} & 72.23 (68,646)         & \textbf{29.44 (25,302)} & 26.96 (73,275)         & \textbf{80.47  (26,880)} & 71.00 (76,683) \\
                Weighted          & \textbf{97.96 (26,880)} & 93.46 (68,646)         & \textbf{75.65 (25,302)} & 51.13 (73,275)         & \textbf{98.11 (26,880)}  & 93.00 (76,683)\\
                \bottomrule
                \end{tabularx}
                \end{table*}

\subsubsection{BiLSTM}

As an alternative method for sequence classification of pages, we use a bidirectional long short-term (biLSTM)~\cite{hochreiter1997long} layer to capture sequential dependencies at the feature extraction level---as opposed to the FM+CRF baseline, which only does so at the prediction level. We experiment with two different kinds of input: the activations of the first FC layer of the FM (128-dimensional vectors); and the concatenation of the pre-computed image and text embeddings (7,936-dimensional vectors), obtained as described in Section~\ref{subSec:methodsFusion}.

The network consists of a biLSTM layer with 128 units for each direction followed by batch normalisation, dropout and an FC layer. When using concatenated image and text embeddings as input, we first apply batch normalisation and dropout, followed by an FC layer with 512 units.

We train four model variants: 
\begin{enumerate}
  \item BiLSTM, which uses fusion activations as input;
  \item BiLSTM-CRF, with the same input and a CRF head on top of the described network;
  \item BiLSTM-F, which uses concatenated image and text embeddings as input; and 
  \item BiLSTM-F-CRF, with the same input and a CRF head.
\end{enumerate} 
Due to the memory footprint of BiLSTM-F and BiLSTM-F-CRF, we use mini-batches of eight lawsuits when training them. All models are trained for one cycle~\cite{cyclelr} of 20 epochs. We use the range test to choose learning rates.

\section{Results and Discussion}
\label{sec:disc}
Table~\ref{table:finalResults} exhibits the \fone~scores of the best performing models, categorised by whether they use textual (CNN), visual (ResNet50-w and ResNet50), textual and visual (FM), or sequential (FM+CRF and BiLSTM-F) information.

All models beat majority class classifiers considering both weighted and average \fone~scores---except for ResNet50-w, whose weighted \fone~score is 25.98 p.p. lower. The models with textual data performed much better than those with only visual information available, which is not surprising given that text content is more discriminative than visual aspects when considering the dataset documents---most of them are similar white pages with blocks of text (Figure \ref{fig:firstPages}).

Regarding fusion and sequence classification results, it is clear that each additional information source contributed to classification metrics: considering average \fone~scores, the FM surpassed the CNN by 0.89 p.p., while the FM+CRF and BiLSTM-F beat the FM by 1.19 and 1.53 p.p., respectively.

In the paragraphs below we will further examine the results of each category (text, image, fusion and sequence) and perform an ablation analysis of the Fusion Module.

\subsection{Text Classification Results}
\label{subSec:resultsText}
Table~\ref{table:textComp} compares the validation performance of our approaches for text classification. Using class frequency penalty weights to help with data imbalance did not help: the CNN-w average and weighted scores were 12.90 and 3.39 p.p. lower than its counterpart with no such strategy. Despite using the same architecture, we achieved better results than the ones reported by Luz de Araujo et al~\cite{luzDeAraujo_etal_VICTOR_LREC_2020}. This is probably because we save model parameters only on validation metric improvement when training.

\subsection{Image Classification Results}
\label{subSec:resultsImage}

Table~\ref{table:imgComp} compares the validation performance of the image classification models. 

The ResNet50 achieved higher average and weighted scores than its counterpart that uses class frequency penalty weights. That said, since the ResNet50 scores for \textit{Acórdão}, ARE and \textit{Despacho} were zero or close to zero, the ResNet50-w scores were more equally distributed across the different classes. With the intuition that this could lead to more discriminative features, we experiment with both models' activations when fusing textual and visual data.

\subsection{Image and Text Combination Results}
\label{subSec:resultsFusion}
Table~\ref{table:FMcomp} shows the performance of the FM trained for 10 epochs with different hyperparameter configurations. Using learnable embeddings for missing textual or visual data proved to be fundamental, improving average \fone~scores by 6.47 and 2.75 p.p. for the models with 512 and 128 hidden units, respectively. While the smaller model performed best, we hypothesise that with further parameter tuning and longer training the bigger model would surpass it.

Table~\ref{table:zeroImgAct} compares the scores of the alternative fusion approaches with the ones from the FM. All of them performed much worse, with decreases in average \fone~score ranging from 2.16 to 14.52 p.p. These results signal how the increase in performance seen by the FM is due to the fusion of data sources; not to different training conditions or model capacity---combining visual and textual data helps.


\subsection{Sequence Classification}
\label{subSec:resultsSequence}

Table~\ref{table:seqComp} compares the validation performance of the LSTM models. To ensure a fair comparison to the other approaches, though we use IOB tagging scheme during training, when reporting results we consider only the original classes. If a given page is an ARE, for example, the predictions B-ARE and I-ARE would both be considered correct, regardless of the position of the page in its lawsuit.

The variants that use as input the image and text embeddings (BiLSTM-F and BiLSTM-F-CRF) outperformed the ones that use the FM activations (BiLSTM and BiLSTM-CRF). This suggests that it is beneficial to jointly learn how to consider sequential dependencies and how to combine multi-modal information.
Surprisingly, the CRF layer helped the BiLSTM model, with an increase in 1.29/0.25 average/weighted \fone~scores, but not the BiLSTM-F model. This may be an artifact of our training settings, with its limited number of training epochs.



    \subsection{First page evaluation}
    \label{subSec:firstPage}
    Table~\ref{table:pageEval} shows the difference in classification performance of samples that are the first page of a document versus those that are interior pages, considering all levels of data availability (text, image, and fusion). 
    
    The first page sample set obtained average/weighted \fone~scores 5.54/4.50, 2.48/24.52 and 9.47/5.11 p.p. higher than its complement, for the text, image and fusion levels, respectively. These results confirm our hypothesis that the first pages are more informative from the point of view of both textual and visual data. Therefore, one possible improvement for page classification of the legal documents is training under a multi-task setting that jointly learns to classify pages and establish document boundaries.

\section{Conclusion}
\label{sec:conc}
In this paper, we presented a novel dataset of Brazilian lawsuits with visual and textual data and proposed a method for sequence-aware multimodal classification of pages from legal documents. Our proposed Fusion Module combines visual and textual features extracted from convolutional neural networks trained separately on image and text data. We experiment with two approaches for sequence classification: post-processing the predictions of the Fusion Module using a linear-chain conditional random field; and training bidirectional LSTM models that alternatively use as input Fusion Module activations or the concatenation of image and text embeddings. Our Fusion Module outperformed the unimodal models, with an ablation analysis confirming that improvement is due to the combination of modalities. We find that learning embeddings for missing visual or textual input is much better than using a vector of zeroes for such cases. Sequence classification of pages brought further improvements, with the best performing model jointly learning how to combine modalities and consider sequential dependencies.

Therefore, future work would include the end-to-end training of the full pipeline: image and text feature extractors, Fusion Module and sequence modelling. Moreover, it is worthwhile to explore if transformer-based~\cite{vaswani_etal_transformer_nips_2017} text encoders such as BERT~\cite{bert} and T5~\cite{raffel_etal_t5_JMLR_2020} can further improve classification performance.


%
\section*{Declarations}
\subsection*{Funding}
This study was financed in part by the Coordenação de Aperfeiçoamento de Pessoal de Nível Superior - Brasil (CAPES) - Finance Code 001. TdC received support
from Conselho Nacional de Desenvolvimento Científico e Tecnológico (CNPq), grant
PQ 314154/2018-3. We acknowledge the support of ``Projeto de Pesquisa \& Desenvolvimento de aprendizado de máquina (machine learning) sobre dados judiciais das repercussões gerais do Supremo Tribunal Federal - STF''. We are also grateful for the support from Fundação de Apoio à Pesquisa do Distrito Federal (FAPDF, project KnEDLe,
convênio 07/2019) and Fundação de Empreendimentos Científicos e Tecnológicos (Finatec). TdC is currently on a leave of absence from the University of Brasilia and works at Vicon Motion Systems, Oxford Metrics Group.

\subsection*{Conflicts of interest}
The authors declare that they have no conflict of interest.

\subsection*{Availability of data}
Data used in this work is available at \url{http://ailab.unb.br/victor/
lrec2020/}.

\subsection*{Code availability}
Code used in this work is available at \url{https://github.com/peluz/victor-visual-text}.

\bibliographystyle{spmpsci}      
\bibliography{mybib}   

\begin{thebibliography}{10}
\providecommand{\url}[1]{{#1}}
\providecommand{\urlprefix}{URL }
\expandafter\ifx\csname urlstyle\endcsname\relax
  \providecommand{\doi}[1]{DOI~\discretionary{}{}{}#1}\else
  \providecommand{\doi}{DOI~\discretionary{}{}{}\begingroup
  \urlstyle{rm}\Url}\fi

\bibitem{audebert_etal_multiModalDocs_arxiv_2019}
Audebert, N., Herold, C., Slimani, K., Vidal, C.: Multimodal deep networks for
  text and image-based document classification.
\newblock CoRR \textbf{abs/1907.06370} (2019).
\newblock \urlprefix\url{http://arxiv.org/abs/1907.06370}

\bibitem{bojanowski_etal_fasttext_arxiv_2016}
Bojanowski, P., Grave, E., Joulin, A., Mikolov, T.: Enriching word vectors with
  subword information.
\newblock arXiv preprint arXiv:1607.04606  (2016)

\bibitem{buitinck_etal_sklearn_ECML_2013}
Buitinck, L., Louppe, G., Blondel, M., Pedregosa, F., Mueller, A., Grisel, O.,
  Niculae, V., Prettenhofer, P., Gramfort, A., Grobler, J., Layton, R.,
  VanderPlas, J., Joly, A., Holt, B., Varoquaux, G.: {API} design for machine
  learning software: experiences from the scikit-learn project.
\newblock In: ECML PKDD Workshop: Languages for Data Mining and Machine
  Learning, pp. 108--122 (2013)

\bibitem{chen_and_blostein_diaSurvey_IJDAR_2007}
Chen, N., Blostein, D.: A survey of document image classification: problem
  statement, classifier architecture and performance evaluation.
\newblock International Journal of Document Analysis and Recognition (IJDAR)
  \textbf{10}(1), 1--16 (2007).
\newblock \doi{10.1007/s10032-006-0020-2}.
\newblock \urlprefix\url{https://doi.org/10.1007/s10032-006-0020-2}

\bibitem{veryDeepTextClf}
Conneau, A., Schwenk, H., Barrault, L., Lecun, Y.: Very deep convolutional
  networks for text classification.
\newblock In: Proceedings of the 15th Conference of the European Chapter of the
  Association for Computational Linguistics: Volume 1, Long Papers, pp.
  1107--1116. Association for Computational Linguistics, Valencia, Spain
  (2017).
\newblock \urlprefix\url{http://www.aclweb.org/anthology/E17-1104}

\bibitem{lsa}
Deerwester, S., Dumais, S.T., Furnas, G.W., Landauer, T.K., Harshman, R.:
  Indexing by latent semantic analysis.
\newblock Journal of The American Society for Information Science
  \textbf{41}(6), 391--407 (1990)

\bibitem{bert}
Devlin, J., Chang, M., Lee, K., Toutanova, K.: {BERT:} pre-training of deep
  bidirectional transformers for language understanding.
\newblock CoRR \textbf{abs/1810.04805} (2018).
\newblock \urlprefix\url{http://arxiv.org/abs/1810.04805}

\bibitem{engin_etal_multiModalBank_ICAIMM_2019}
Engin, D., Emekligil, E., Oral, B., Arslan, S., Akp{\i}nar, M.: Multimodal deep
  neural networks for banking document classification.
\newblock In: International Conference on Advances in Information Mining and
  Management, pp. 21--25 (2019)

\bibitem{he_etal_resnet_CVPR_2016}
{He}, K., {Zhang}, X., {Ren}, S., {Sun}, J.: Deep residual learning for image
  recognition.
\newblock In: IEEE Conference on Computer Vision and Pattern Recognition
  (CVPR), pp. 770--778 (2016).
\newblock \doi{10.1109/CVPR.2016.90}

\bibitem{hochreiter1997long}
Hochreiter, S., Schmidhuber, J.: Long short-term memory.
\newblock Neural computation \textbf{9}(8), 1735--1780 (1997)

\bibitem{ulmfit}
Howard, J., Ruder, S.: Universal language model fine-tuning for text
  classification.
\newblock In: Proceedings of the 56th Annual Meeting of the Association for
  Computational Linguistics (Volume 1: Long Papers), pp. 328--339. Association
  for Computational Linguistics, Melbourne, Australia (2018).
\newblock \doi{10.18653/v1/P18-1031}.
\newblock \urlprefix\url{https://www.aclweb.org/anthology/P18-1031}

\bibitem{batchnorm}
Ioffe, S., Szegedy, C.: Batch normalization: Accelerating deep network training
  by reducing internal covariate shift.
\newblock In: Proceedings of the 32nd International Conference on Machine
  Learning - Volume 37, pp. 448--456. JMLR.org (2015).
\newblock \urlprefix\url{http://proceedings.mlr.press/v37/ioffe15.html}

\bibitem{jain_et_wigington_ICDAR_2019}
{Jain}, R., {Wigington}, C.: Multimodal document image classification.
\newblock In: International Conference on Document Analysis and Recognition
  (ICDAR), pp. 71--77 (2019).
\newblock \doi{10.1109/ICDAR.2019.00021}

\bibitem{kingma2014adam}
Kingma, D.P., Ba, J.: Adam: A method for stochastic optmisation.
\newblock In: International Conference on Learning Representations ({ICLR})
  (2015).
\newblock Preprint available at \url{https://arxiv.org/abs/1412.6980}

\bibitem{CRF}
Lafferty, J.D., Andrew, M., Pereira, F.C.N.: Conditional random fields:
  Probabilistic models for segmenting and labeling sequence data.
\newblock In: Proceedings of the Eighteenth International Conference on Machine
  Learning, ICML, pp. 282--289. Morgan Kaufmann Publishers Inc., San Francisco,
  CA, USA (2001)

\bibitem{luzDeAraujo_etal_VICTOR_LREC_2020}
{Luz de Araujo}, P.H., de~Campos, T.E., Ataides~Braz, F., Correia~da Silva, N.:
  {VICTOR}: a dataset for {B}razilian legal documents classification.
\newblock In: Proceedings of The 12th Language Resources and Evaluation
  Conference (LREC), pp. 1449--1458. European Language Resources Association,
  Marseille, France (2020).
\newblock \urlprefix\url{https://www.aclweb.org/anthology/2020.lrec-1.181}

\bibitem{mota_etal_cnnPet_ENIAC_2020}
Mota, C., Lima, A., Nascimento, A., Miranda, P., de~Mello, R.: Classificação
  de páginas de petições iniciais utilizando redes neurais convolucionais
  multimodais.
\newblock In: Anais do XVII Encontro Nacional de Inteligência Artificial e
  Computacional, pp. 318--329. SBC, Porto Alegre, RS, Brasil (2020).
\newblock \doi{10.5753/eniac.2020.12139}.
\newblock
  \urlprefix\url{https://sol.sbc.org.br/index.php/eniac/article/view/12139}

\bibitem{raffel_etal_t5_JMLR_2020}
Raffel, C., Shazeer, N., Roberts, A., Lee, K., Narang, S., Matena, M., Zhou,
  Y., Li, W., Liu, P.J.: Exploring the limits of transfer learning with a
  unified text-to-text transformer.
\newblock Journal of Machine Learning Research \textbf{21}(140), 1--67 (2020).
\newblock \urlprefix\url{http://jmlr.org/papers/v21/20-074.html}

\bibitem{IOB_scheme}
Ramshaw, L.A., Marcus, M.P.: Text chunking using transformation-based learning.
\newblock In: Natural language processing using very large corpora, pp.
  157--176. Springer (1999).
\newblock \doi{10.1007/978-94-017-2390-9_10}.
\newblock Preprint available at \url{http://arxiv.org/abs/cmp-lg/9505040}

\bibitem{rusinol_etal_multimodalAdmDocument_IJDAR_2014}
Rusi{\~{n}}ol, M., Frinken, V., Karatzas, D., Bagdanov, A.D., Llad{\'o}s, J.:
  Multimodal page classification in administrative document image streams.
\newblock International Journal on Document Analysis and Recognition (IJDAR)
  \textbf{17}(4), 331--341 (2014).
\newblock \doi{10.1007/s10032-014-0225-8}.
\newblock \urlprefix\url{https://doi.org/10.1007/s10032-014-0225-8}

\bibitem{russakovsky_etal_imagenet_IJCV_2015}
Russakovsky, O., Deng, J., Su, H., Krause, J., Satheesh, S., Ma, S., Huang, Z.,
  Karpathy, A., Khosla, A., Bernstein, M., Berg, A.C., Fei-Fei, L.: {ImageNet
  Large Scale Visual Recognition Challenge}.
\newblock International Journal of Computer Vision (IJCV) \textbf{115}(3),
  211--252 (2015).
\newblock \doi{10.1007/s11263-015-0816-y}

\bibitem{sandler_etal_mobileNetV2_CVPR_2018}
Sandler, M., Howard, A., Zhu, M., Zhmoginov, A., Chen, L.C.: Mobilenetv2:
  Inverted residuals and linear bottlenecks.
\newblock In: Proceedings of the IEEE Conference on Computer Vision and Pattern
  Recognition (CVPR) (2018)

\bibitem{justiceSummary2020}
{Secretaria de Comunicação Social do Conselho Nacional de Justiça}: Sumário
  executivo do relatório justiça em números 2020 (2018).
\newblock
  \urlprefix\url{https://www.cnj.jus.br/wp-content/uploads/2020/08/WEB_V2_SUMARIO_EXECUTIVO_CNJ_JN2020.pdf}

\bibitem{simonyan_etal_vgg16_2015}
Simonyan, K., Zisserman, A.: Very deep convolutional networks for large-scale
  image recognition.
\newblock In: International Conference on Learning Representations (2015)

\bibitem{lrfinder}
{Smith}, L.N.: Cyclical learning rates for training neural networks.
\newblock In: IEEE Winter Conference on Applications of Computer Vision (WACV),
  pp. 464--472 (2017).
\newblock \doi{10.1109/WACV.2017.58}

\bibitem{cyclelr}
Smith, L.N., Topin, N.: Super-convergence: Very fast training of neural
  networks using large learning rates.
\newblock CoRR \textbf{abs/1708.07120} (2017).
\newblock \urlprefix\url{http://arxiv.org/abs/1708.07120}

\bibitem{smith_Tesseract_icdar2007}
Smith, R.: An overview of the {Tesseract} {OCR} engine.
\newblock In: Ninth International Conference on Document Analysis and
  Recognition ({ICDAR}), vol.~2, pp. 629--633. {IEEE} (2007)

\bibitem{stf_victor_news}
{Supremo Tribunal Federal}: {Ministra Cármen Lúcia anuncia início de
  funcionamento do Projeto Victor, de inteligência artificial} (2018).
\newblock
  \urlprefix\url{http://www.stf.jus.br/portal/cms/verNoticiaDetalhe.asp?idConteudo=388443}

\bibitem{vaswani_etal_transformer_nips_2017}
Vaswani, A., Shazeer, N., Parmar, N., Uszkoreit, J., Jones, L., Gomez, A.N.,
  Kaiser, L.u., Polosukhin, I.: Attention is all you need.
\newblock In: I.~Guyon, U.V. Luxburg, S.~Bengio, H.~Wallach, R.~Fergus,
  S.~Vishwanathan, R.~Garnett (eds.) Advances in Neural Information Processing
  Systems 30, pp. 5998--6008. Curran Associates, Inc. (2017).
\newblock
  \urlprefix\url{http://papers.nips.cc/paper/7181-attention-is-all-you-need.pdf}

\bibitem{wiedemann_heyer_multiModalPSS_LRE_2019}
Wiedemann, G., Heyer, G.: Multi-modal page stream segmentation with
  convolutional neural networks.
\newblock Language Resources and Evaluation  (2019).
\newblock \doi{10.1007/s10579-019-09476-2}

\end{thebibliography}

\end{document}